%% file: ms.tex
\documentclass[letterpaper, 10 pt, conference]{ieeeconf}  %

\IEEEoverridecommandlockouts                              %

\usepackage{latexsym}
\usepackage{graphicx}
\usepackage{subfig}
\usepackage{amsmath}
\usepackage{amssymb}
\usepackage{blindtext}
\usepackage{todonotes}
\usepackage{wrapfig}
\graphicspath{{figures/}}
\let\labelindent\relax
\usepackage{enumitem, kantlipsum}
\usepackage{caption}
\usepackage{multirow}
\usepackage{diagbox}
\usepackage[ruled]{algorithm2e}
\usepackage{makecell}
\usepackage{tabularx}
\usepackage{rotating}
\usepackage{titlesec}
\usepackage[hidelinks]{hyperref}
\usepackage{xcolor}
\titlespacing* %
    {\subsection}
    {0.5mm}
    {1.5ex plus 1ex minus .1ex} %
    {1.3ex plus .1ex} %
\definecolor{velvet}{HTML}{E37222}
\hypersetup{
  colorlinks   = true, %
  urlcolor     = velvet, %
  linkcolor    = velvet, %
  citecolor   = velvet %
}

\usepackage{romannum}
\title{\LARGE
 \bf
AcTExplore: Active Tactile Exploration on Unknown Objects
}
\newcommand{\AlgName}{AcTExplore}

\author{Amir-Hossein Shahidzadeh, Seong Jong Yoo, Pavan Mantripragada,\\ Chahat Deep Singh, Cornelia Ferm\"{u}ller, Yiannis Aloimonos\thanks{All authors are associated with the Perception and Robotics Group at the University of Maryland, College Park. The support by Brin Family Foundation, the Northrop Grumman Mission Systems University Research Program, ONR under grant award N00014-17-1-2622 and National Science Foundation under grant BCS 1824198 are gratefully acknowledged.}
\thanks{Supplementary material is available at \href{http://prg.cs.umd.edu/AcTExplore}{{\color{blue}http://prg.cs.umd.edu/AcTExplore}}.}
}

\begin{document}

\maketitle
\thispagestyle{empty}
\pagestyle{empty}

\setlength{\parindent}{5pt}

\begin{abstract}
Tactile exploration plays a crucial role in understanding object structures for fundamental robotics tasks such as grasping and manipulation. However, efficiently exploring such objects using tactile sensors is challenging, primarily due to the large-scale unknown environments and limited sensing coverage of these sensors. To this end, we present \AlgName, an active tactile exploration method driven by reinforcement learning for object reconstruction at scales that automatically explores the object surfaces in a limited number of steps. Through sufficient exploration, our algorithm incrementally collects tactile data and reconstructs 3D shapes of the objects as well, which can serve as a representation for higher-level downstream tasks. Our method achieves an average of \textbf{95.97}\% IoU coverage on unseen YCB objects while just being trained on primitive shapes.

\end{abstract}

\section{Introduction}

\label{sec:introduction}

Human perception of the environment is a multifaceted process that involves multi-sensor modalities, including vision, audition, haptic, and \textit{proprioception}. While deep learning has made significant progress in visual perception, conventional vision-only models have limitations compared to human perception. Humans excel at perceiving objects in challenging environments, utilizing their multi-sensor inputs \cite{munozMultisensoryPerceptionUncertain2012, cappeSelectiveIntegrationAuditoryvisual2009} such as the eyes for visual properties and the skin for tactile sensing which is essential to characterize physical properties such as texture, stiffness, temperature, and contour \cite{https://doi.org/10.1002/adma.202203073, doi:10.1098/rsos.181563, doi:10.1098/rstb.2011.0171}. Thus, vision and tactile sensation have distinct roles in scene perception, each with unique requirements. Vision relies on direct line-of-sight unobstructed views, whereas tactile sensation only necessitates physical contact, enabling perception in challenging scenarios like occluded or dark environments when vision is limited. This distinction underscores the value of tactile sensing in scene perception, motivating the development of \AlgName{}. Our goal is to maximize contact with the object's surface during exploration, thereby fully utilizing the benefits of tactile sensation that aligns with the capabilities of humans and other living beings to bridge the gap between machine and human perception.

The human skin, our largest organ, allows us to perceive contact with the external world. This has prompted behavioral studies \cite{JOHANSSON198217, trends, Massari} which investigate human manipulation skills, where the significance of tactile sensing becomes evident \cite{yuan2023robot}. Similarly, the successful automation of robotic manipulation tasks heavily relies on their perceptual capabilities. Consequently, a multitude of tactile sensors have been developed for robotic applications, encompassing optical-based sensors like DIGIT \cite{DBLP:journals/corr/abs-2005-14679}, GelSight \cite{s17122762} and Soft-bubble \cite{soft-bubble}, which demonstrate remarkable proficiency in discerning skin deformation \cite{suresh2022midastouch}. Conversely, bio-inspired electrode-based sensors such as SynTouch BioTac \cite{biotac} necessitate extensive post-processing and finite-element modeling to accurately represent skin deformation, as studied  in \cite{DBLP:journals/corr/abs-2101-05452, DBLP:journals/corr/abs-2103-16747}.

\begin{figure}[t!]
    \centering
    \includegraphics[width=0.99\columnwidth]{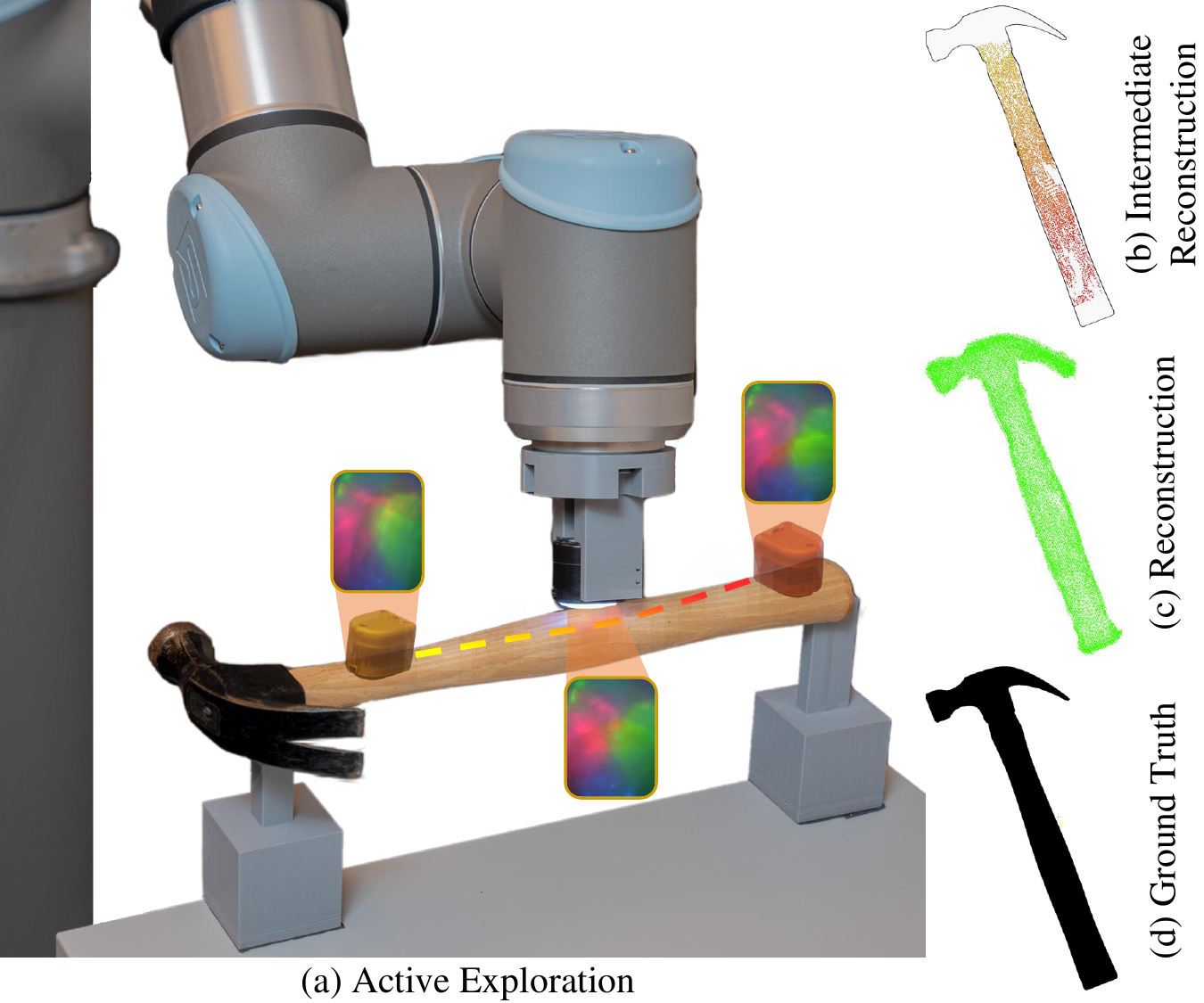}
    \caption{\textbf{Reconstruction of a hammer}. (a) showcases the trajectory of the tactile sensor in 3D space. (b) depicts the intermediate tactile readings on the hammer's surface, with the color gradient representing the passage of time. Following thorough tactile exploration, we achieve a complete object reconstruction (c), highlighting the effectiveness of our active strategy in exploring the entire reachable surface.}
    \label{fig:banner}
    \vspace{-0.6cm}
\end{figure}

Tactile sensors output a detailed local perspective of objects, giving them a unique role in surface exploration \cite{suresh2023neuralfeels, caddeo2023sim2real, 10410896}. However, predicting future actions (moves) based on a single touch to explore an entire object's shape is challenging. It involves two main issues: the agent getting stuck in cyclical movements or revisiting explored areas and the difficulty in inferring which action will maintain touch to avoid exploring empty space while consistently exploring the object. Balancing exploration and maintaining contact requires an active policy considering the tactile reading history and sensor trajectory. Our active policy facilitates the sensor to avoid \emph{Non-Exploratory Scenarios}, enabling us to explore the objects within a limited number of actions. In \textbf{\AlgName}, we address the aforementioned challenges by formulating it as a Partially Observable Markov Decision Process (POMDP) where the policy is computed from only a recent trajectory rather than the full interaction history.

To this end, we propose \textbf{\AlgName}, an active method for tactile exploration that utilizes deep reinforcement learning, and the \textbf{core contributions} are given as follows: \textbf{(a)} \emph{Exploring Object's Surface} with minimal actions without limiting the approach to a specific distribution of objects. We achieve this by training the agent to learn dexterous movements from fundamental actions on primitive shapes (cube, sphere, etc.). Remarkably, as demonstrated in the experiments (Sec. \ref{sec:exp}), the learned behavior extends to unseen objects. \textbf{(b)} Introducing \emph{Temporal Tactile Sensing} in the state representation (Sec. \ref{method:staterep}) to enable Short-Term Memory (STM) on taxels (tactile receptors), inspired by various neurological and behavioral studies \cite{LAWSON2015239, KATUS2015275}. \textbf{(c)} Proposing a curiosity-driven \emph{Active Exploration Algorithm} for 3D reconstruction at scale that can be integrated into various high-level tasks in the future, such as Grasp Pose Refinement \cite{DBLP:journals/corr/abs-2103-00655}, Scene Perception \cite{9562061,Suresh22icra}.

\section{Related Works}
\label{sec:citations}
\textbf{Active SLAM}
involves traversing an unknown environment while simultaneously localizing and constructing a map
~\cite{chenSelflearningExplorationMapping2019, curiosity, chaplotLearningExploreUsing2020, chenAutonomousExplorationUncertainty2020,liang2019salientdso}. In traditional SLAM, revisiting the marked areas is beneficial for correcting estimated localization errors~\cite{SLAMReview, ActiveSLAMReview}. However, we aim to achieve an efficient exploration pipeline that minimizes revisits. Active SLAM is generally formulated as a Partially Observable Markov Decision Process (POMDP) ~\cite{ActiveSLAMReview}, with various reward function formulations, including curiosity~\cite{curiosity}, coverage~\cite{chenLearningExplorationPolicies2019}, or entropy-based~\cite{botteghiREINFORCEMENTLEARNINGHELPS2020}. 
Unlike the conventional active SLAM setting, which operates in unknown 2D spaces, our work focuses on exploring a 3D workspace with limited sensing space, leading to ambiguity \cite{Tac2Pose}. 

\textbf{Deep reinforcement learning in exploration.}
Recent advances in computing power and physics-based simulators have boosted research in virtual navigation and exploration.
For instance, \cite{DBLP:journals/corr/MirowskiPVSBBDG16} trained an Asynchronous Advantage Actor-Critic (A3C) agent in a 3D maze, incorporating long short-term memory (LSTM) to provide the memory capability \cite{Ramani2019ASS, DBLP:journals/corr/abs-1712-03316}.
Furthermore, \cite{placedDeepReinforcementLearning2020} tackled the robot exploration problem using the D-optimality criterion as an intrinsic reward, significantly accelerating the training process. In our problem, we incorporate an exploration bonus as a reward function. This approach incentivizes the agent to explore undiscovered state and action pairs, leading to more sample efficient algorithms ~\cite{azar2017minimax, jin2018qlearning}.   

\textbf{Tactile applications.} Tactile information plays a crucial role in human perception, encompassing tasks from object manipulation to emotional expressions. As a result, tactile sensors have been employed in various applications~\cite{linDTactVisionBasedTactile2022, suresh2022midastouch, parkBiomimeticElastomericRobot2022,ganguly2020grasping2,ganguly2022gradtac}. 
Especially when robots manipulate deformable objects, tactile sensors provide meaningful information that enhances system robustness alongside vision sensors~\cite{pecynaVisualTactileMultimodalityFollowing2022}. 
Furthermore, tactile information has been used to estimate the pose of the objects~\cite{10160359, Tac2Pose, sureshTactileSLAMRealtime2021, zhao2023fingerslam} or the relative pose of the gripper for object handling ~\cite{kelestemurTactilePoseEstimation2022}. 
Similar to our work, tactile sensors have been employed to identify or reconstruct unknown objects~\cite{7177729, linLearningIdentifyObject2019, hoganSeeingYourSkin2021, OttenhausActiveExploration}.
For instance, \cite{xu2023tandem3d} designed a tactile object classification pipeline that actively collects tactile information while exploring the object. 
The closest previous studies that address the 3D reconstruction of unknown objects through information-theoretic exploration were ~\cite{luCuriosityDrivenSelfsupervised2022, 8793773, 7759723}. Some of these were evaluated in simulation only, and some on simple objects very different from the real world YCB \cite{DBLP:journals/corr/CalliWSSAD15} objects, which present challenges in collision avoidance during planning when the objects' geometry is unknown.
Some other studies have focused on shape reconstruction, specifically handling missed segments individually \cite{DBLP:journals/corr/abs-2007-03778, DBLP:journals/corr/abs-2201-01367}. 
However, using a passive exploration algorithm, their primary focus was on shape completion. 
In contrast, our work addresses the challenges of 3D object active exploration in both simulation and real-world, facilitating the reconstruction process by exploring the object in limited trials.

\begin{figure}%
    \centering
    \includegraphics[width=\columnwidth]{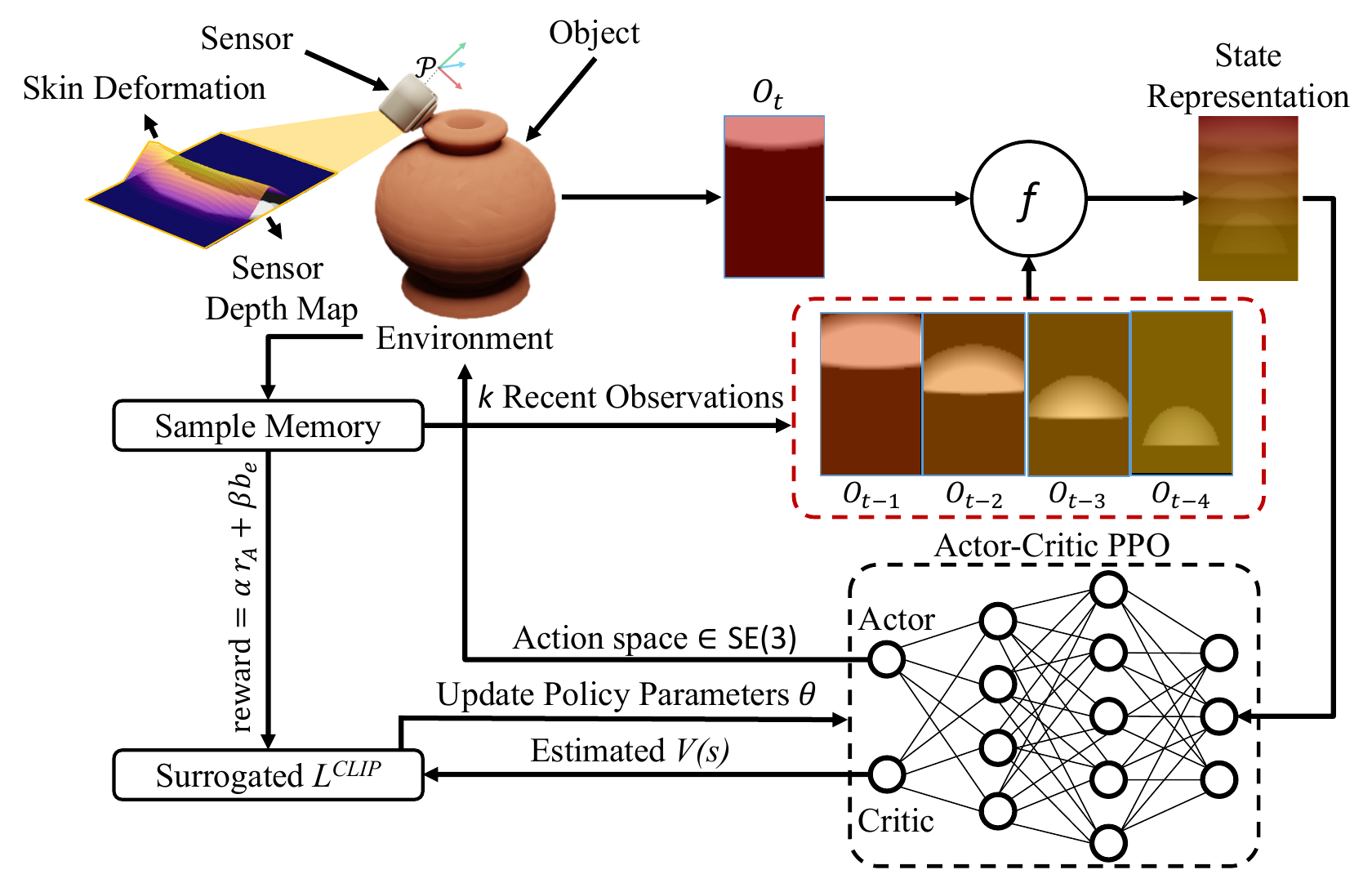}
    \caption{\textbf{Overview.} This figure illustrates the key steps and components of \AlgName\ in a scenario where the sensor moves upward along the jar's edge. We employed Temporal Tactile Averaging for state representation $f$ (Sec. \ref{method:staterep}) to encode consecutive observations, enabling the perception  of movement on the sensor vital for learning dexterous actions. We also incorporate an Upper Confidence Bound (UCB) exploration as a bonus to encourage effective exploration.}%
    \label{fig:overview}
    \vspace{-0.6cm}
\end{figure}

\section{Method}
\label{sec:method}
In \AlgName, we consider a tactile sensor mounted on a robotic arm end-effector interacting with an unknown fixed 3D object. We have access to the end-effector pose $\mathcal{P}_t \in SE(3)$ from the forward kinematics (FK). The goal is to navigate an unknown object's entire surface within a limited workspace to collect tactile data for the reasons mentioned in Sec. \ref{sec:introduction}. At time $t$, the model will utilize consecutive tactile data $\{O_{t}, O_{t-1}, \ldots, O_{t-(k-1)}\}$ to generate exploratory action $a_t \in SE(3)$ based on $k$ recent observations. The effectiveness of the Completeness process heavily relies on the robot's exploration algorithm.

An exploration policy, denoted as $\pi_\theta$ (referred to as the explorer) determines the next action, $\arg\max_{a_{t}} \pi_\theta(a_{t}|s_t)$,  based on state $s_t$ to maximize the cumulative reward (Sec. \ref{method: reward}) which serves as an estimate of area coverage over an unknown object's surface. Therefore, the problem formulation of \AlgName\, encompasses four key components: 

\subsection{Observation Space}
At each time step $t$, the sensor observes skin deformation as an image (Fig. \ref{fig:overview}) that can be converted into a depth map of skin~\cite{DBLP:journals/corr/abs-2005-14679, suresh2022midastouch}. We'll denote this depth image as observation space $O_t \in R^{H \times W}$, corresponding to the sensor's deformation at pose $\mathcal{P}_t$.

\subsection{State Representation}
\label{method:staterep}
The state serves as the sole input to the explorer, so it has to be sufficiently informative, enabling the model to generate exploratory actions. 
Let $s_t$ be the state input for $\pi_\theta$ at time $t$. Considering the possibility that multiple optimal actions correspond to the observation $O_t$ at finger pose $\mathcal{P}_t$, it is advantageous to construct a state representation $s_t$ that incorporates short-term memory. This can be achieved by using a sequence of $k$ consecutive observations $\{O_t, O_{t-1},\ldots, O_{t-(k-1)}\}$ to encapsulate the complexity of the state, enabling the policy to generate the appropriate action for long-horizon exploration. Therefore, $s_t = f(O_t, O_{t-1},\ldots, O_{t-(k-1)})$, where $f$ can be any function representing spatio-temporal information on the sensor like optical flow, which is however computationally costly to generate on the fly. This function aims to reduce the dimensionality of $s_t$ while extracting critical features for the state representation \cite{doi:10.1177/0278364919887447}, which are not feasible to infer from a single observation. This is particularly valuable for learning high-level, complex actions that require a wider view. We conduct extensive experiments investigating the effectiveness of various state representations in Sec. \ref{sec:exp}. In Fig. \ref{fig:overview}, we visualize the resulting tactile readings and our proposed functions $f$:
\begin{equation}
     \text{TTS}(O_t, O_{t-1}, ..., O_{t-(k-1)}):  O_t \mathbin\Vert O_{t-1} \mathbin\Vert \ldots \mathbin\Vert O_{t-(k-1)}
     \label{eq:TTS}
     \tag{Temporal Tactile Stacking}
\end{equation}\vspace{-10pt}

\begin{equation}
     \text{TTA}(O_t, O_{t-1}, ..., O_{t-(k-1)}): \sum_{i=0}^{k-1} \
     \alpha_i O_{t-i}
     \label{eq:TTA}
     \tag{Temporal Tactile Averaging}
\end{equation}
\vspace{0.1cm} 
where $\alpha_i$ are decreasing and $\sum_{i=0}^{k-1} \alpha_i = 1$ so the most recent reading $O_{t}$ is the most effective observation in $s_t$, and others will be averaged respectively.

\begin{figure}
\centering
\includegraphics[width=0.95\columnwidth]{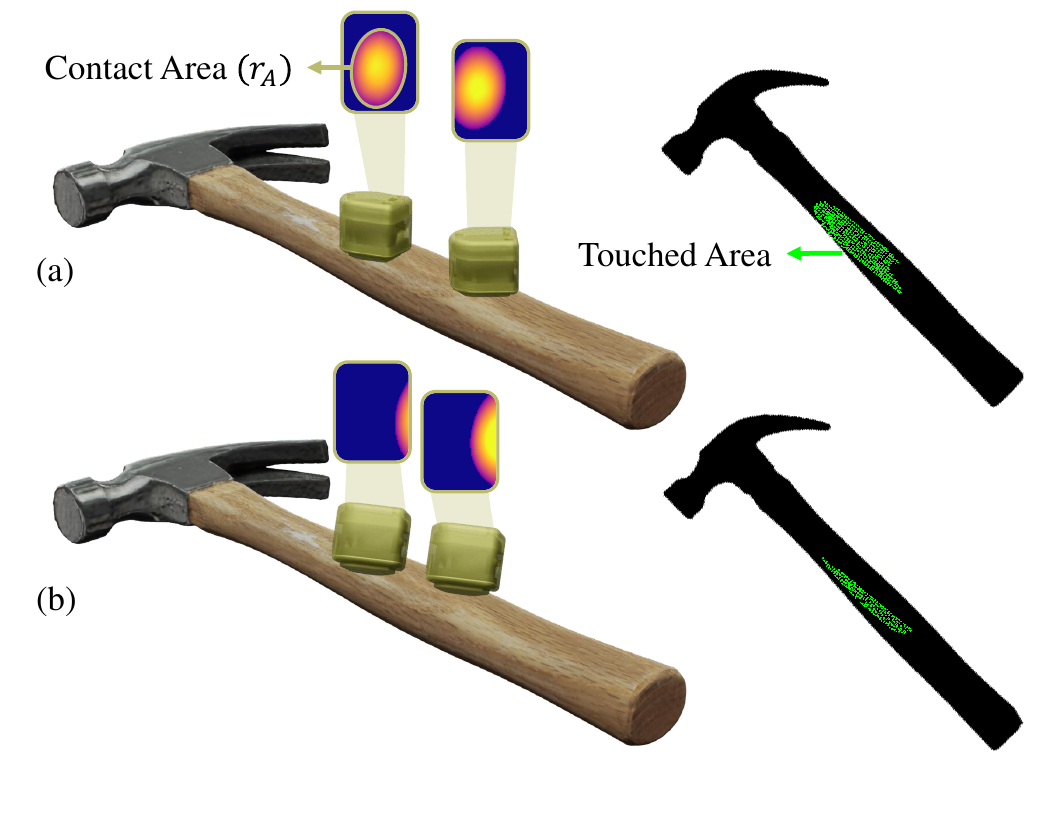}
\captionof{figure}{ \textbf{Depth readings sliding over hammer} (a): Sensor aligned with object's surface, receiving more depth information and moving stably. (b): Misaligned rotation increases the probability of losing contact in future steps.}
\label{fig:rewArea}
\vspace{-1em}
\end{figure}

\subsection{Action Space}
To efficiently explore the complex geometry of 3D objects, we enable the finger to move in a 6-degree-of-freedom (6DOF) action space denoted as $A \in SE(3)$. In this action space, we consider small incremental translations $(x, y, z)$ and rotations ($\gamma$, $\theta$, $\psi$) around the workspace frame, with the bottom of the finger as the reference point. The model selects one of the dimensions $(x, y, z, \gamma, \theta, \psi)$ and either increases or decreases its value by the specified step. This results in a total of 12 possible actions within the action space. This action space is chosen to facilitate the control of the sensor under kinematic and collision constraints of the arm in both simulation and real world.

While this action space can facilitate exploration and interaction, it can be enhanced by adding an action that allows the finger to return to the last touching location. This additional action serves the purpose of touch recovery and addressing Non-Exploratory scenarios where contact with the object may be temporarily lost. By including this \emph{Touch Recovery action} ($a_{TR}$), the robot can reestablish contact and give up on the trajectories that aren't worthy of further exploration. It will also guarantee that the model never strays too far from the object, as it learns to perform this action after a certain number of steps without any touch. Therefore, the action space consists of $12+1$ actions in total.
\subsection{Reward}
\label{method: reward}
We can break down the exploration objective into easier sub-goals. Specifically, in this work, we want to maximize objects' coverage through the tactile sensor's trajectory. To accomplish this, the reward function is divided into two components:

\begin{enumerate}
[wide, labelwidth=!, labelindent=0pt]

\item \textbf{Contact Area Reward}
measures the contact area between the robot's finger and the object's surface. The rationale is that a larger contact area corresponds to more information being gathered from the object, as illustrated in Fig. \ref{fig:rewArea}. In other words, this reward will encourage actions that align the tactile sensing area of the finger with the object's surface. 

\item \textbf{Exploration Bonus} is intended to encourage the agent further to explore the workspace from a global perspective. Drawing inspiration from \cite{azar2017minimax,jin2018qlearning}, we have introduced a memory mechanism to track all the agent's trajectories. This memory allows us to keep a count of the number of times the agent has performed a particular action $a$ in a specific pose $\mathcal{P}$, denoted as $N(\mathcal{P},a)$. 
Even so, using $N(\mathcal{P},a)$ directly for exploration is not practical because most of the states would have $N(\mathcal{P},a)=0$, especially considering that the workspace is often continuous or high-dimensional. So instead, we define $\hat{N}(\mathcal{P},a)$ as the the number of close poses (Sec. S\Romannum{1}-C-1$^{\ref{suppFN}}$) in the trajectory history.
By having access to $\hat{N}(\mathcal{P},a)$ for each pose-action pair, we can incorporate a bonus term $b_e = \frac{1}{\sqrt{\hat{N}(\mathcal{P},a)}}$ into the area reward $r_A$ in case that $r_A>0$ which indicates that the sensor is touching the object. 
The bonus term $b_e$ is a term that incentivizes exploration and curiosity throughout the trajectory. By incorporating this bonus, actions that have been infrequently taken in the past on a specific $\mathcal{P}$ are rewarded more prominently, encouraging the agent to explore less-visited state-action pairs.
Furthermore, the agent is guided toward more comprehensive exploration and exhibits a tendency to venture into uncharted regions of the workspace. This aids in mitigating the impact of sub-optimal local optima and fosters a broader understanding of the environment, resulting in improved convergence and enhanced exploration behavior as discussed in \cite{jin2018qlearning}. 

\item \textbf{Penalties} 
The agent might learn two possible trivial local optima scenarios to maximize the area reward and exploration bonus without exploring the object in long-horizon. To address this issue, we'll define the necessary negative rewards:
\textbf{Revisit Penalty ($P_{rev}$) :} To prevent the agent from learning policies that involve revisiting recently visited poses, we'll construct a short-term memory $\mathcal{D}=\{\mathcal{P}_{i} \}_{i=t-m}^{t}$ of $m$ recent interactions by time $t$. If an action leads the sensor to revisit a pose already present in its short-term history $\mathcal{D}$, a revisit penalty $P_{rev}<0$ will be imposed to discourage such trajectories.
\textbf{Touch Recovery Penalty ($P_{TR}$):} One possible scenario that may not be covered by $P_{rev}$ is when the sensor moves freely in space without making contact with the object for more than $m$ steps ($|\mathcal{D}|$), and then performs a touch recovery action, which has a positive area reward ($r_A>0$). To prevent such scenarios, we introduce a negative reward $P_{TR}$ each time the agent selects the touch recovery action ($a_{T R}$). However, despite this negative reward, the agent still tends to choose the touch recovery action due to the high value of $V(s_{t+1})$ associated with recovery actions. Additionally, we can control the number of actions without touch by adjusting  $P_{TR}$.    
\end{enumerate}

By penalizing non-exploratory scenarios, the agent is incentivized to explore new areas, mitigating the risk of getting stuck in sub-optimal loops and performing dexterous actions on a long horizon (further discussion in Sec. S\Romannum{1}-B-1\footnote{Supplementary material: \href{http://prg.cs.umd.edu/AcTExplore}{{\color{blue}http://prg.cs.umd.edu/AcTExplore}} \label{suppFN}}).

With all of these considerations in mind, the reward function is formulated as follows:
\begin{equation}
    r(s_t,a_t) = 
    \begin{cases}
        \alpha r_A + \frac{\beta}{\sqrt{\hat{N}(\mathcal{P}_t,a_t)}}, & \text{if } r_A > 0 \text{ and } \mathcal{P}_{t+1} \notin \mathcal{D}    \\
        P_{rev}, & \text{if } r_A > 0 \text{ and } \mathcal{P}_{t+1} \in \mathcal{D} \\
        P_{TR} , & \text{if } a_t = a_{TR}\text{\small{(touch recovery)}} \\
        0, & \text{otherwise} \\
    \end{cases}
\end{equation}

where $r_A$ represents the reward based on the area of contact between the sensor and the object's surface at time $t$, $P_{rev}$ denotes the penalty term applied to actions leading to a previously visited pose in $\mathcal{D}$ and $\hat{N}(\mathcal{P}_t,a_t)$ signifies number of times the agent has performed action $a_t$ in pose $\mathcal{P}_t$ by time $t$, which is used to calculate the exploration bonus term.

By combining these components within the reward function,  we aim to achieve a balance between contact area maximization (i.e., \emph{exploitation}), and avoidance of non-exploratory scenarios (i.e., \emph{exploration}) which addresses the mentioned difficulties for training in unknown environments.

To compute variance-reduced advantage/value function estimators, \AlgName{} utilizes a modified Proximal Policy Optimization (PPO) algorithm by modeling the exploration objective as an intrinsic auxiliary reward while enriching the state with temporal representation. We summarize our overall method in Alg. \ref{alg:overall}. The environment is initialized with an object to be explored and a tactile sensor to move on the object's surface and collect observations actively. We define the exploration workspace such that the first touch with the object happens as the sensor moves toward the \textbf{workspace's center}. The agent continues to interact with the object to learn optimal actions through the multi-objective reward \textbf{AMB} (Sec. \ref{method: reward}) that estimates coverage that is not available during training on unknown objects.

\begin{algorithm}
    \caption{\textsc{\AlgName{}, Procedure}}\label{alg:overall}
    \For{episode = \text{1,2, ...} }
        {
        $\mathcal{D} \gets$ List of size $m$ \\
        $\mathcal{P}_0 \gets $ First touch pose\\
        $O_0 \gets $ Tactile sensor reading at $\mathcal{P}_0$\\ 
        $\hat{N}(p,a) \gets 0$ for all $(p,a) \in \mathcal{P} \times \mathcal{A}$\\ 
            \For{t = 0, 1, 2, \ldots, T-1}
            {
                $s_t \gets f(\{O_i\}_{i=t-\min(k,t)}^t)$\\
                $a_t \gets \arg\max_{a'} \pi_{\theta}(a'|s_t)$ \\ 
                $\mathcal{P}_{t+1}, O_{t+1} \gets$ tactileSensor.step($a_t$) \\
                $\hat{N}(\mathcal{P}_{t}, a_t) \gets \hat{N}_t(\mathcal{P}_{t}, a_t) + 1 $ \\
                $r_A \gets $ nonZeroCount($O_t$) $/$ size($O_t$) \\
                $r \gets 0$ \\
                \uIf{$a_t = a_{TR}$} { 
                   $r \gets P_{TR}$ \hfill \tcp{\small{touch recovery}}
                }
                \uElseIf{$\mathcal{P}_{t+1} \in \mathcal{D}$}{
                    $r \gets P_{rev}$ \hfill \tcp{\small{revisit}}
                }
                \uElseIf{$r_A > 0$}{
                    $b_e \gets \frac{1}{\sqrt{\hat{N}_t(\mathcal{P}_t, a_t)}}$ \\
                    $r \gets \alpha r_A + \beta b_e $\\
                }
                $\mathcal{D}$.add($\mathcal{P}_{t+1}$) \\
                $\delta_t \gets r + \gamma V(s_{t+1})  - V(s_t)$ 
                \\
            }
        Compute advantages $\hat{A}_{i \in [T]}: \sum_{j=i}^{T-1} \gamma^{j-i} \delta_{i+j}$ \\
        $\theta \gets$ Optimize surrogate $L^{CLIP}(\theta, \hat{A}_{[T]})$
        } 
\end{algorithm}

\begin{figure}[t!]
    \centering
    \includegraphics[width=\columnwidth]{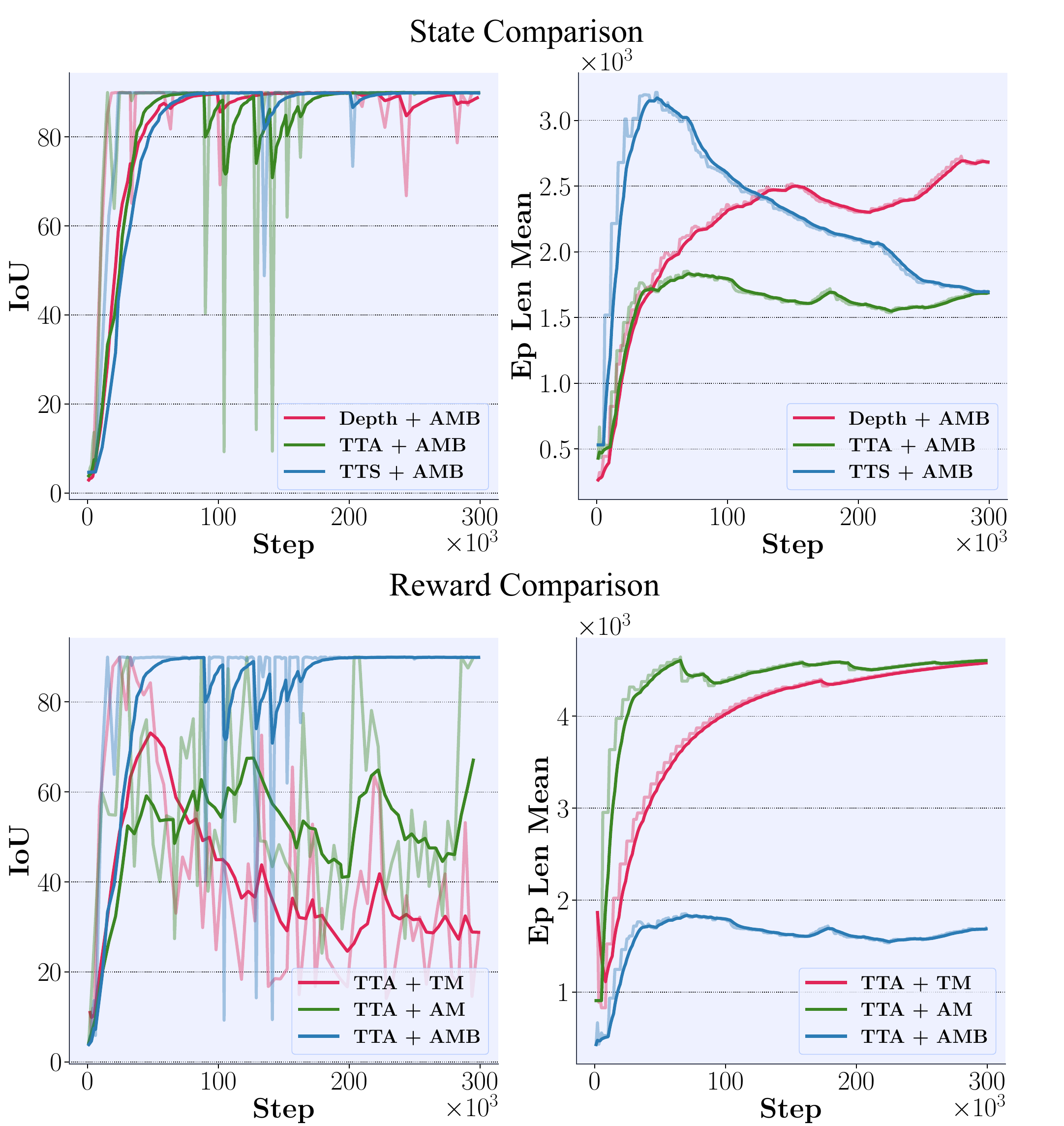}
    \caption{\textbf{Training results.} For each row: The \textbf{[top]}  compares state representations using AMB for the reward function, while the \textbf{[bottom]} showcases different reward settings using TTA for state representation. Note that episodes terminate when the agent surpasses 90\% IoU, reaches the horizon steps, or reaches the workspace boundaries.}
    \label{fig:training_plots}
\end{figure}

\section{Experiments}
\label{sec:exp}

This section evaluates and analyzes our method \AlgName{}, with various rewards and states on zero-shot (\textit{unseen}) objects. In addition, we validate our method on over 400 quantitative and qualitative experiments in reconstructing unknown objects with varying complexities. In our experiments, \emph{we use reconstruction accuracy as the metric for tactile exploration with a limited number of steps as it represents the \AlgName\, exploration potential}.

\subsection{Experimental Setup}
\textbf{Simulation.} We employ TACTO \cite{DBLP:journals/corr/abs-2012-08456, pybullet} to simulate tactile sensor skin deformation during object interactions and modified PPO from StableBaselines3 \cite{stable-baselines3} in \AlgName{}.

The TACTO simulator is calibrated with real-sensor data to ensure Sim-to-Real generalization. It generates depth map images from real-world signals, serving as our observation $O$.
We train the agent only with primitive objects -- sphere and cube -- for 300K steps. These primitive objects are selected as they represent a broad range of object shapes, with the sphere having curvature and the cube having sharp edges, corners, and flat surfaces. To assess the model's performance, we evaluate it on YCB objects that were not encountered during training time. This evaluation demonstrates the efficacy of training with primitives, which exhibit strong generalization capabilities for objects with realistic textures (Fig. S2$^{\ref{suppFN}}$). For the termination condition, each episode either spans 5000 steps (Sec. S\Romannum{1}-A$^{\ref{suppFN}}$) or concludes once the Intersection over Union (IoU) metric exceeds 90\%, or when the sensor leaves the workspace boundaries. This strategy is adopted to reduce the training time. In Tab. \ref{tab:evaluation}, we show that these termination conditions do not limit the IoU performance during testing as our methods achieved over 90\%.

\begin{figure}[ht!]
    \centering
\includegraphics[width=0.98\columnwidth]{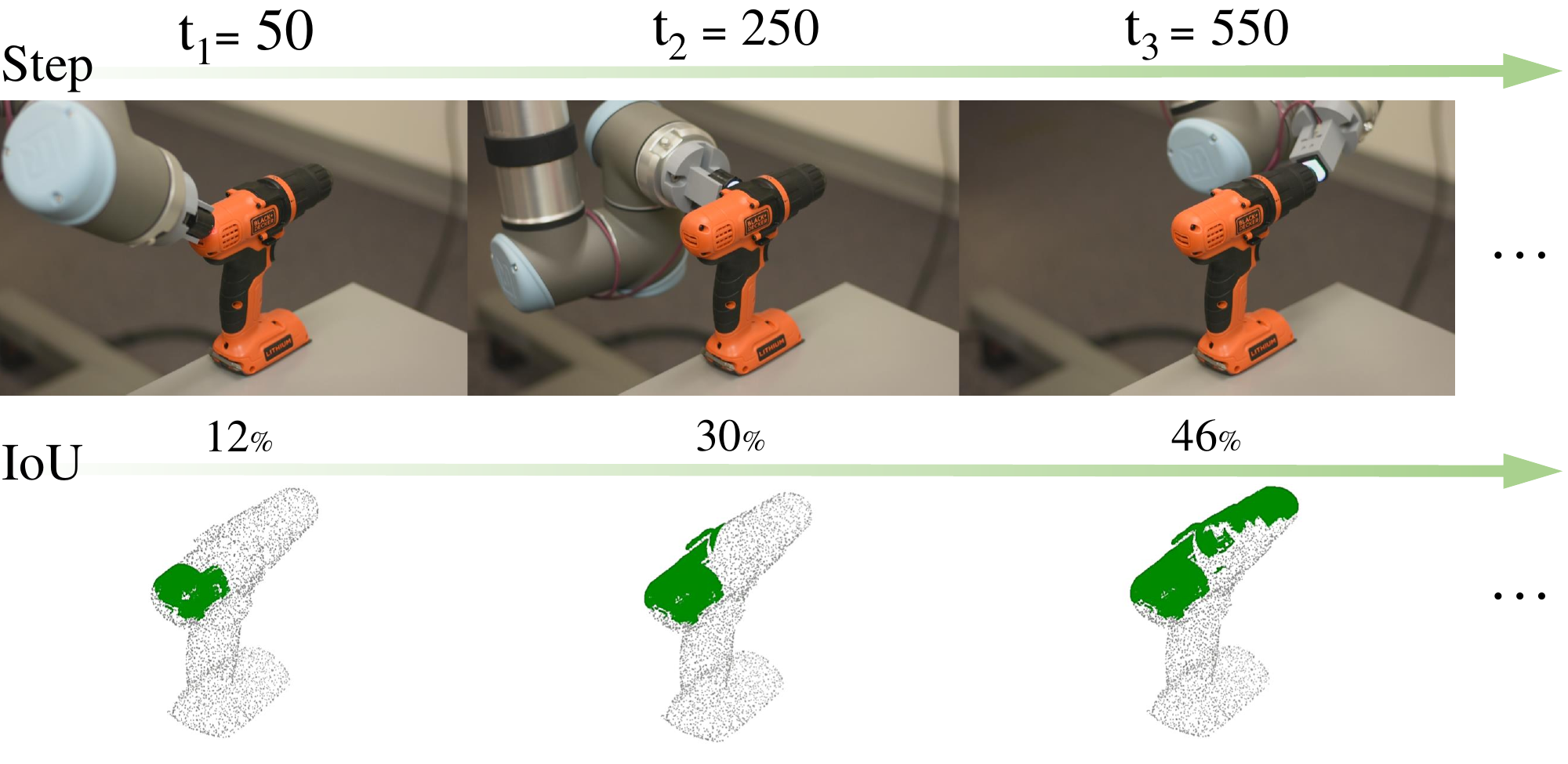}
\caption{\textbf{Real-World Exploration Execution.}  Still frames from \AlgName{}'s exploration of a drill, starting from the rear and progressing towards the chuck. The second row shows the covered area per step, with IoU computed over the exploration workspace above the drill's grip. $t_i$ is the $i$-th step of the trajectory.}
\label{fig:drill-real-world}
\end{figure}

\textbf{Real-World System.} We employ a UR10 arm to manipulate the 6D pose of the DIGIT (Fig. \ref{fig:drill-real-world}). This control is achieved by transforming changes in the DIGIT’s frame into a set of joint trajectories via inverse kinematics which are facilitated with ur\_rtde. The resulting trajectories are executed only if free from self-collisions and within the defined workspace. When an invalid trajectory is generated, we select an alternative action based on the PPO's advantage values.

Unlike simulations, where consecutive action executions while in contact with the object have minimal impact, our real-world implementation introduces significant shearing on the sensing surface. To ensure the safe execution of actions generated by our policy, we have adopted a strategy of lifting the DIGIT in the normal direction of the contact after each contact event. This strategy remains well-founded due to our policy's consistent alignment of our sensor surface with the object's surface and does not compromise its general applicability. Our method successfully transferred to real-world experiments without requiring further fine-tuning. Fig. \ref{fig:drill-real-world} illustrates the effectiveness of our exploration policy on a drill in the real-world.

\textbf{Baselines Configuration.} To evaluate the efficacy of each component, we have established a collection of baselines for three different state rep. and reward functions in Tab. \ref{tab:baselines}. 

\vspace{-5pt}
\begin{table}[ht]
    \centering
    \renewcommand{\arraystretch}{1.5}
    \caption{Baseline Formulations. \textbf{TTA}: Temporal Tactile Averaging, \textbf{TTS}: Temporal Tactile Stacking (concatenation is denoted as $\mathbin\Vert$ ), \textbf{TM}: binary Touch indicator ($\mathbb{I}(\cdot)$) + short Memory, \textbf{AM}: contact Area + short Memory, \textbf{AMB}: contact Area + short Memory + UCB Bonus}
    \label{tab:baselines}
    \begin{tabular}{ll|c|c|c}
    
    \Xhline{1pt}
    \multicolumn{2}{c|}{\multirow{2}{*}{State}} & Depth & TTA & TTS \\ 
    \cline{3-5}
    & & $O_t$ & $\sum_{i=0}^{k-1}\alpha_i O_{t-i}$ &$O_t \mathbin\Vert \ldots \mathbin\Vert O_{t-(k-1)}$ \\
     \Xhline{1pt}
     \multicolumn{2}{c|}{\multirow{2}{*}{Reward}} & TM & AM & AMB \\
     \cline{3-5}
     & & $\mathbb{I}(O_t)$ & $r_A(O_t)$ & $\alpha r_{A}(O_t) + \frac{\beta}{\sqrt{\hat{N}(\mathcal{P}_{t}, a_t)}}$ \\
     \Xhline{1pt}
    
    \end{tabular}
\vspace{-1.2em}
\end{table}

\setlength{\tabcolsep}{0.8em}
{
\begin{table*}[ht]
    \vspace{1mm}
    \centering
    \renewcommand{\arraystretch}{1.1}
    \caption{\textbf{Quantitative results on unseen YCB objects:} The table presents IoU and Chamfer-L1 distance (cm) \cite{Mescheder_2019_CVPR} between ground-truth and predicted meshes from methods in Tab. \ref{tab:baselines}. The surface area is listed below each object's name as a severity metric. The details of metrics, confidence intervals, and step counts are given in the supplementary material$^{\ref{suppFN}}$.}
    \label{tab:evaluation}
    \resizebox{2\columnwidth}{!}{
    \begin{tabular}
    {l|l|c|c|c|c|c|c|c}
    \Xhline{2.5\arrayrulewidth}
        \multicolumn{2}{c|}{\multirow{2}{*}{\textbf{Methods}}} & \textbf{Can}  & \textbf{Banana} & \textbf{Strawberry} & \textbf{Hammer} & \textbf{Drill} & \textbf{Scissors} & \textbf{Mustard} \\ 
        
        \multicolumn{2}{c|}{} &\scriptsize{(616 $\text{cm}^2$)} &\scriptsize{(216 $\text{cm}^2$)} &\scriptsize{(68 $\text{cm}^2$)} &\scriptsize{(410 $\text{cm}^2$)} &\scriptsize{(591 $\text{cm}^2$)} &\scriptsize{(165.48 $\text{cm}^2$)} &\scriptsize{( 454.54 $\text{cm}^2$)} \\ 
        \hline
        \multicolumn{6}{c}{} \scriptsize{IoU $\uparrow$} \hspace{2 em} \scriptsize{(Chamfer-$L_1$ $\downarrow$)} \\
        \Xhline{2.5\arrayrulewidth}
        \multirow{3}{*}{TM} 
        & depth   & 31.93 (2.66) & 11.11 (7.52) & 83.60 (0.44) & 32.78 (1.86)& 19.19 (4.1) & 24.29 (8.15) & 10.07 (4.07)  \\ 
        & TTA  & 17.60 (3.57) & 6.03 (9.03) & 41.0 (1.23) & 14.85 (6.94)& 28.15 (3.99)  &14.17 (4.98) & 19.94 (3.22)  \\ 
         & TTS & 15.93 (5.22) & 18.23 (5.48) & 57.89 (0.88) & 28.66 (2.47)&15.5 (3.97)  & 11.26(4.97) & 14.55(2.95)  \\ 
         \Xhline{2.5\arrayrulewidth}

        \multirow{3}{*}{AM} 
        &depth &11.59 (5.49) &10.22 (6.84)  &47.33 (1.16)  & 5.07 (7.69) &9.49 (4.03)  & 5.11(6.78) & 11.04(5.16)\\ 
        & TTA  & 72.70 (0.56) & 97.70 (0.35) & \textbf{100} (0.28) & 79.80 (0.82) & 57.58 (1.43)  &41.77 (2.87) &71.72 (0.80) \\ 
         & TTS  & \textbf{98.25} (0.22) & \textbf{100} (0.34) & \textbf{100} (0.31) & 88.22 (0.44)& 99.02 (0.37 )  & 28.37 (2.38) & 87.13 (0.59)  \\ 
         \Xhline{2.5\arrayrulewidth}
    
        \multirow{4}{*}{AMB} 
        & depth   & 41.45 (1.42) & 98.64 (\textbf{0.25}) & \textbf{100} \textbf{(0.23)} & 61.42 (1.17) & 79.68 (0.95)  & 31.99 (3.2) & 65.74 (0.9) \\ 
          & depth+LSTM  & 88.54 (0.3) & 99.96 (0.28) & \textbf{100} (0.24) & 87.54 (0.49) & 92.81 (0.36) &  29.83 (0.58) & 88.33 (0.36) \\
        & TTA (ours) & 89.6 (0.29)  & \textbf{100} (0.33) & \textbf{100} (0.25) & \textbf{98.22} (0.29)& 98.85 (0.32)  &  67.02 (0.87) & \textbf{95.91} (0.51) \\ 
         & TTS (ours) & 97.45 \textbf{(0.20)} & \textbf{100} (0.3) & \textbf{100} (0.25) & 96.96 \textbf{(0.28)} & \textbf{99.74 (0.31)}  & \textbf{74.62} (\textbf{0.61}) & 95.02 (\textbf{0.49})\\ 
         \Xhline{2.5\arrayrulewidth}
    \end{tabular}
    }
    \vspace{-6mm}
\end{table*}
}

\begin{figure}[b!]
    \vspace{-0.4cm}
    \centering
    \includegraphics[width=\linewidth]{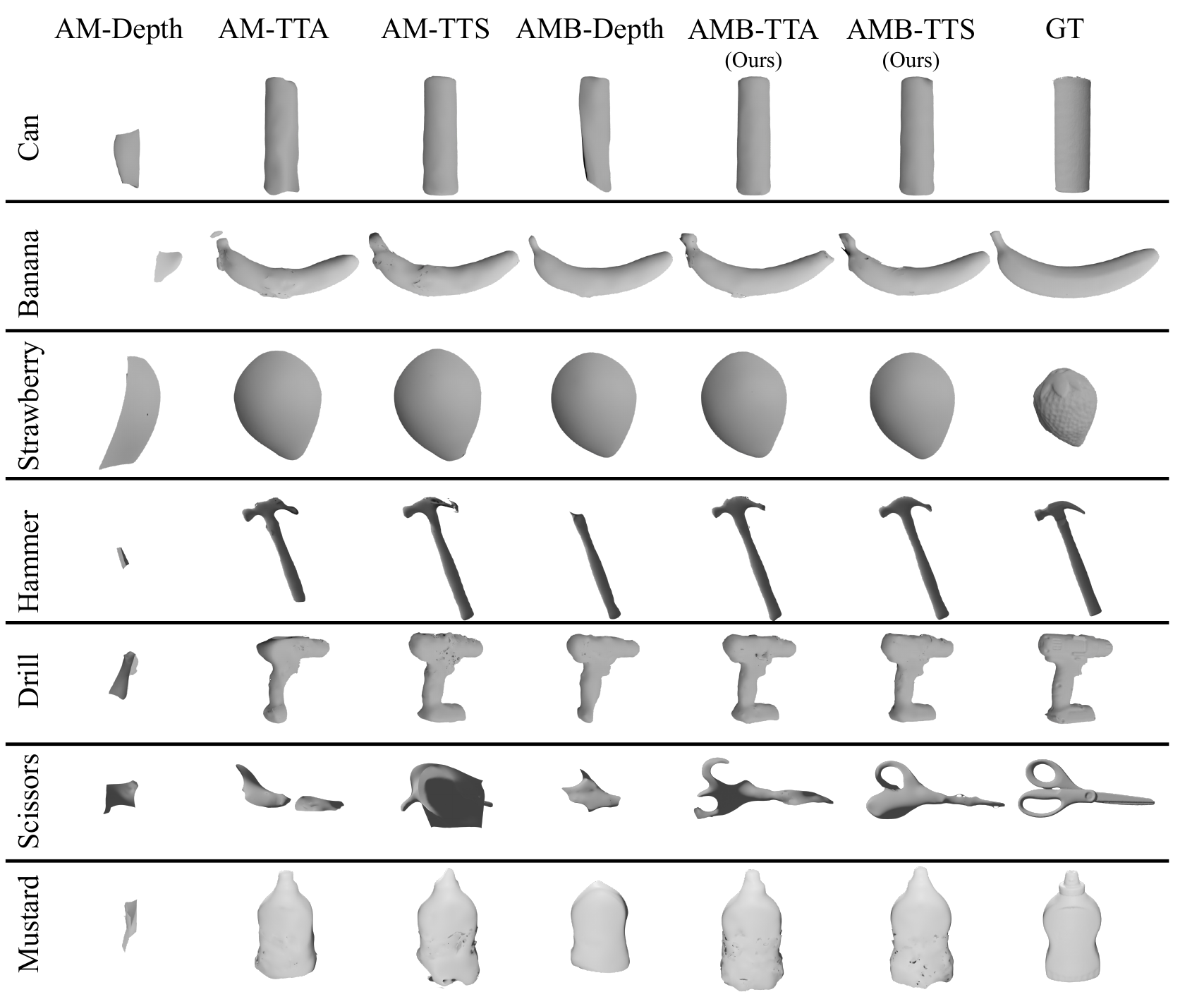}
    \caption{\textbf{Qualitative results on unseen YCB objects} with different state and reward settings. We obtain point cloud data from active tactile exploration on the object's surface. To generate a mesh from the collected point cloud, we apply Poisson surface reconstruction algorithm~\cite{kazhdan2006poisson}. Further experiments are provided in supplementary materials.}
    \label{fig:mesh_result}
\end{figure}

\subsection{Analysis \& Discussion}
\textbf{State Comparison.} We compare different state representations using the same reward function (AMB), considering both representations with and without temporal information. This analysis highlights the influential impact of temporal information on learning dexterous and high-level actions. As shown in Fig. \ref{fig:training_plots}, all state representations achieve a specified IoU during training. However, the state representations incorporating temporal information demonstrate higher stability, consistently reaching the 90\% IoU objective after 200K steps. In contrast, the depth-only representation struggles to maintain the IoU objective and is outperformed by temporal representations in Tab. \ref{tab:evaluation}. Furthermore, when considering the number of steps required to achieve the IoU objective, TTS training takes longer than TTA as $s_t^{TTS} \in R^{k \times H \times W}$ is $k$ times bigger than $s_t^{TTA} \in R^{H \times W}$ which is averaging observations rather than stacking them. However, in our experiments in Fig. \ref{fig:mesh_result}, we witnessed that both TTA and TTS are competitive, with TTS excelling on longer objects and TTA performing better on complex shapes.

\textbf{Reward Comparison.}
In our pursuit of efficient exploration, we tried various reward functions mentioned in Tab. \ref{tab:baselines}. During training, we plotted the IoU and episode length until termination in Fig. \ref{fig:training_plots}. Notably, the AMB reward function outperformed the others, satisfying the IoU objective through encouraging exploration of less visited poses. In contrast, TM and AM cannot use environmental feedback as much as AMB can. This limitation arises from TM and AM's deprivation of long-horizon history, which hampers their capacity to gather sufficient information through intrinsic rewards. As a result, AMB is better equipped to leverage environment feedback $\left(\frac{1}{\hat{N}(\mathcal{P},a)}\right)$ effectively for improved exploration and sample efficiency. However, AM outperforms TM as it utilizes contact area information and can still align the sensor's sensing area with the object surface to collect more information and maintain a reliable touch for future actions. Indeed, the disparity between TM and AM can also be understood as the distinction between using a touch sensor versus a tactile sensor for exploring an object.

\textbf{Limitations and Future Work.}
The current formulation of our method has certain limitations. First, it assumes a moving sensor relative to a \textbf{fixed-pose} rigid object, necessitating a physically accurate simulator to narrow the sim2real gap for moving objects. Second, Although \AlgName{} is not restricted by object shape, it is designed to keep the sensor close to recent touching poses. This could pose challenges in environments with disconnected components. Workspace splitting can be a potential solution to address this problem. Third, the sensor exhibits a small depth bias in the simulation resulting in larger reconstructions. While generally negligible, this bias becomes dominant when handling objects roughly the same size as the sensor, such as the strawberry shown in Fig. \ref{fig:mesh_result}. 

As a step towards benchmarking in tactile exploration, we have released our extensive explorations for YCB objects in Tab. S1$^\text{\ref{suppFN}}$ with a maximum of 5000 steps. While employing tactile sensors on multi-finger robotic hands may streamline the exploration process \cite{suresh2023neuralfeels}, there remains a promising direction for future research in modifying the POMDP that effectively handles collisions between sensors while maintaining object generalization.

\vspace{-0.6em}
\section{Conclusion}
\vspace{-0.2em}
In this work, we introduced a novel reinforcement learning method using tactile sensing to explore unknown 3D objects actively. It addresses the need for an active exploration method to enable numerous works \cite{DBLP:journals/corr/abs-2103-00655, 9562061, Suresh22icra, Comi2023TouchSDFAD} to become fully automated. \AlgName\, is not limited to specific shape distributions as it has only been trained on primitive shapes to learn fundamental movements by leveraging temporal tactile information and intrinsic exploration bonuses. We demonstrated this through our experiments with various shape complexities like a drill or a clay pot in both the real world and simulation.

\bibliography{ms}{}
\bibliographystyle{plain}

\clearpage
\newpage

\input{supplement}

\end{document}

%% file: supplement.tex






\begin{figure}[b]
    \centering
    \includegraphics[width=\linewidth]{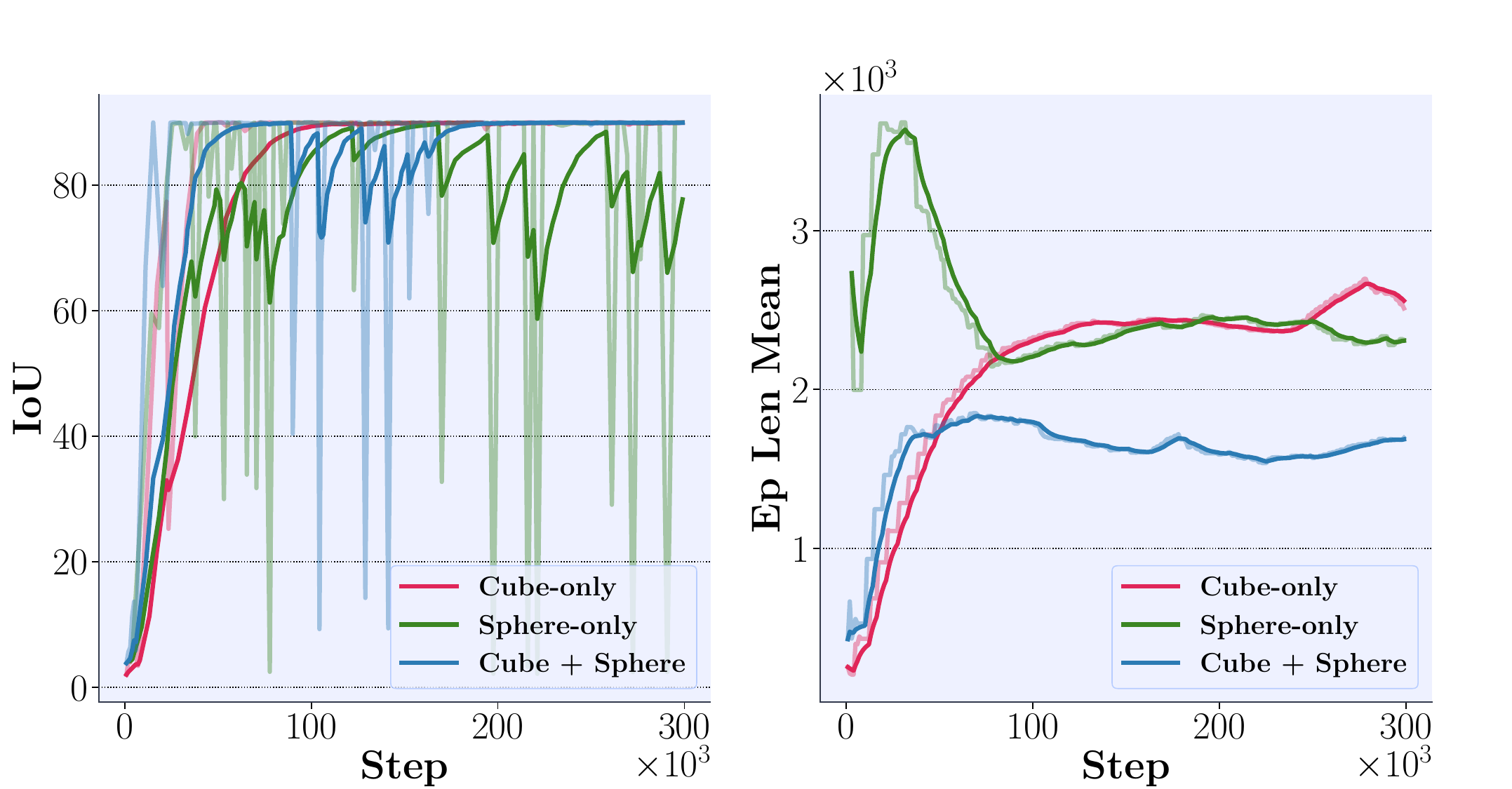}
    \caption{\textbf{Ablation study of training primitives:} We trained AMB-TTS with cube-only and sphere-only setting as well.}
    \label{fig:ablation}
\end{figure}
\section{Implementation Details}
\subsection{Action Space}
\label{S:action_space}
Suppose the sensor can freely move in 3D space, then it has a full 6-DOF continuous action space. However, in order to speed up the training process we discretize the 6-DOF actions into small translations ($x, y, z$) and rotations ($\gamma, \theta, \psi$) steps. The agent can pick one of 6-DOF to decrease or increase which either translates or rotates the sensor. Therefore the 12 action space is $A= \{\pm x, \pm y, \pm z, \pm \gamma, \pm \theta, \pm \psi\}$. 
The translation step ($T_s$) is 4$mm$, while the rotation step ($R_s)$ is 15 degrees about each axis. Furthermore, we introduce \textit{Touch Recovery action} ($a_{TR}$) by saving last touch pose $\mathcal{P}_{TR}$. Note that the quantity of steps required to explore objects is contingent upon the translation and orientation step size of our action space. To provide further clarity, let's consider an example. Consider an object with a surface area of 220$cm^2$. Simplifying this object to a square cube with 90\% of the area, each edge's length would be approximately 
5.7$cm$. Given a translation size of 4$mm$, it necessitates about 206 actions for optimal exploration of each facet. A rotation of 15 degrees necessitates 6 actions to transition between facets. Therefore, exploring a cube of 
 theoretically entails 1260 actions, considering our action step size. Now, if we apply this concept to the YCB's banana, which has a comparable surface area(216$cm^2$) but is more intricate than a cube and necessitates additional rotations, the TTS-AMB requires 1631 actions, contrasting with the 1260 actions needed for the cube which seems reasonable when the object is curved and cylindrical and takes more rotation actions.

\subsection{Reward}
\label{supp:reward}
\subsubsection{Hyperparameters Tuning}
Our reward function encompasses several hyperparameters, the effects of which and tuning methodologies are expounded in this section. Please note that $r_A$ and $b_e$ are normalized in range of $[0,1]$, thus for tuning $\alpha$ and $\beta$ which are designed to regularize $r_A$, and $b_e$ respectively, we have tried various values, maintaining constraint $\alpha + \beta = 1$, to ensure $r(s_t, a_t) \in [0,1]$. Our observations revealed large values of $\alpha$ led to learning policies that moves the agent in a loop which is bigger than short-term memory size $|\mathcal{D}|=m$ as it would receive $P_{rev}$ in smaller loops where the required actions are less than $m$. Conversely, when $\beta$ is too large the agent learn policies where the agent failed to align its sensing area with objects. In consideration of these factors and the distributions of $r_A$ and $b_e$, we determined $\alpha=0.15$ and $b_e=0.85$ to effectively address the outlined issues. Regarding the tuning of $P_{rev}$, it is pertinent to note that its magnitude should be substantial enough to prohibit bad scenarios like loop and non-exploratory trajectories. $P_{rev}$ has a direct interplay with $m$ as it applies solely when the new pose $\mathcal{P}_{t+1} \in \mathcal{D}$ so with with an empirically established $m=20$, $P_{rev}=-0.03$ results in the favorable behavior. $P_{TR}$'s role is to discourage the model from selecting the touch recovery action which has a positive reward as it'll touch the object's surface where ($r_A>0)$. Furthermore, it's actually regulating the number of exploratory actions without touch as the agent is sacrificing the positive rewards of touching poses near the current pose for opting to explore surfaces that may not be directly connected or proximate to the previous pose. Finally, by choosing $P_{TR}=-0.2$, all the mentioned issues will be mitigated. To tune $P_{TR}$, we recommend first tuning the other hyperparameters with 12 actions(without touch recovery action) and subsequently determining the appropriate value for $P_{TR}$ based on the complexity of the environment. The TTA representation also requires some regularizer parameters $\alpha_i$ which are generated from  

\begin{figure}[t]
    \centering
    \includegraphics[width=\linewidth]{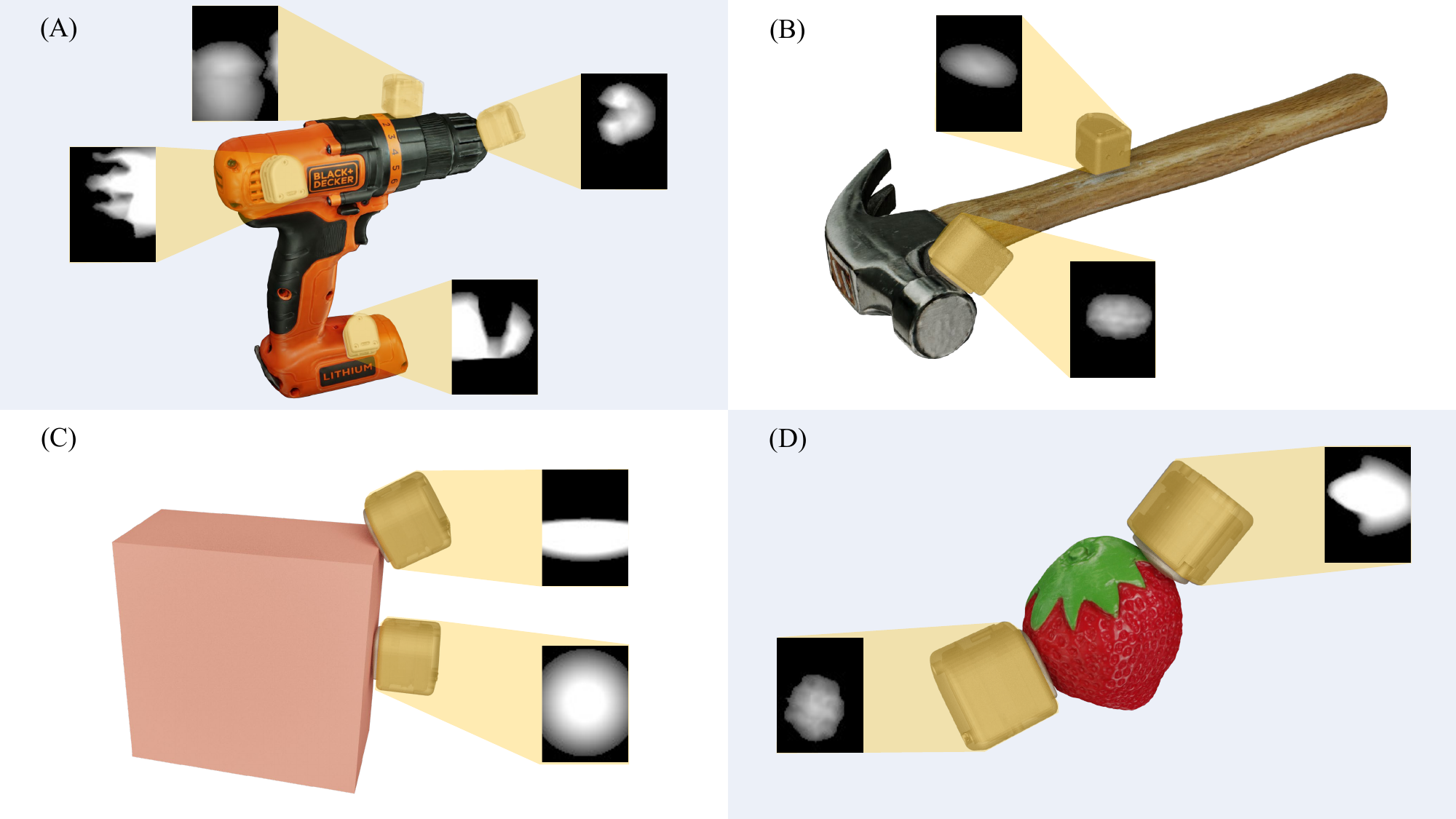}
    \caption{Variety of textures in simulation. (C) One of the primitive objects and tactile depth readings when sensor is touching a flat surface vs an edge. (A, B, D) multiple random poses on some YCB objects and their tactile depth readings, a noticeable distribution shift becomes apparent when comparing plain primitive objects with the real textures on YCB objects. However, Tab.\ref{tab:evaluation}
    indicates that AcTExplore has been generalized enough to adapt to unseen objects.}  
    \label{fig:texture}
\end{figure}

\begin{equation}
    \alpha_i = \frac{1+\frac{i}{\lambda}}{\sum^{m}_{k=0} 1+\frac{k}{\lambda}}
\end{equation}
that satisfies $\sum_{i=0}^{m} \alpha_i = 1$ and will generate the biggest weight for the most recent observation which corresponds to $\alpha_m$. In our experiments, $\lambda=50$ results in the expected behavior from TTA.

\subsection{Further Results}
\begin{figure}[t]
    \centering
    \includegraphics[width=\linewidth]{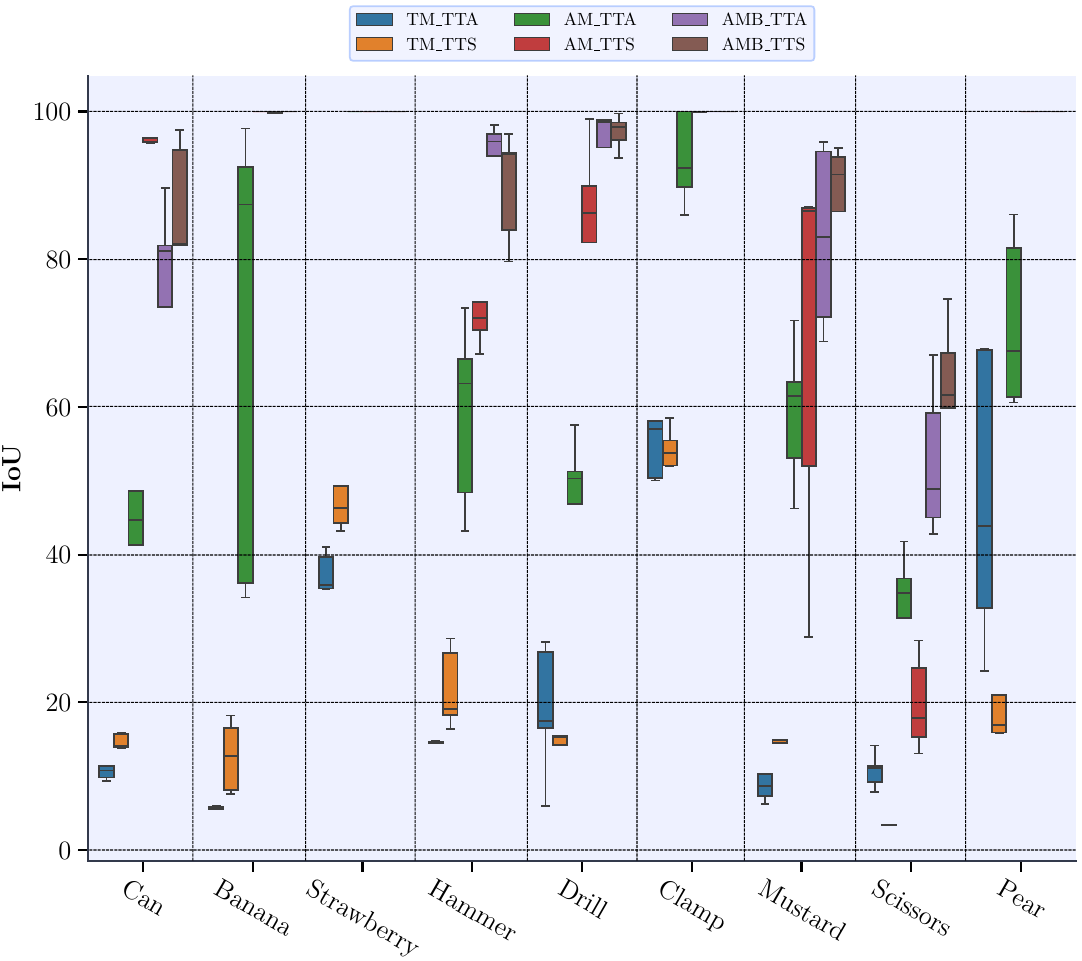}
    \caption{\textbf{Distribution graph:} Since our proposed algorithm is not a deterministic method, we performed 5 trials with each object. Overall, AMB reward function shows small variation and outperform others.}
    \label{fig:bar_graph}
\end{figure}

\subsubsection{Exploration Bonus}
As we have discussed in Sec. \Romannum{3}-D, explicitly defining $N(\mathcal{P},a)$ as the count of times the agent took action $a$ precisely at pose $\mathcal{P}$ throughout the trajectory history is not feasible. This is attributed to the high-dimensionality of the workspace and the likelihood that the agent might not re-encounter pose $\mathcal{P}$. As an alternative approach, we introduced $\hat{N}(\mathcal{P},a)$, denoting the count of times the agent executed action $a$ in proximity to pose $\mathcal{P}$. 

Let's define the sensor's pose as
\begin{equation}
    \mathcal{P}_t = [T_t  |  R_t]
\end{equation}
where $T_t=(x_t,y_t,z_t)$ and $R_t=(\gamma_t, \theta_t, \psi_t)$ is translation and orientation of the sensor at step $t$ respectively. Then $\mathcal{P}_{t'}$ is a close pose to $\mathcal{P}_t$ when it satisfies the following conditions:
\begin{enumerate}
    \item $\|(x_t-x_{t'}, y_t-y_{t'}, z_t-z_{t'})\| \leq trans_{thresh}$
    \item $\arccos(\min(1, \langle R_t, R_{t'} \rangle)) \leq rot_{thresh}$
    \item $a_t=a_{t'}$      
\end{enumerate}
Then we can define
\begin{equation}
\hat{N}(\mathcal{P},a) = \sum_{t=0}^T \mathbb{I}_{close}(\mathcal{P},\mathcal{P}_t).\mathbb{I}(a=a_t) 
\end{equation}

$trans_{thresh}$ and $rot_{thresh}$ needs to be tuned based on sensor's sensing area and translation ($T_s$) and rotation ($R_s$) of action space (Sec. \ref{S:action_space}). In our experiments, we used $trans_{thresh}=2*T_s$ and $rot_{thresh}=4*R_s$.

\subsection{Metrics}
\paragraph{3D Surface IoU}
We introduce 3D surface IoU metric to evaluate our method. We define a set of ground truth point clouds uniformly sampled from target object as $\mathcal{O}^{gt}= \{p_{i}^{gt} \}_{i=1}^{10^5}$ and  $\mathcal{O}^{s}_{t}= \bigcup_{i=1}^t O_i=\{p_{i}^s\}_{i=1}^{tM}$ is the union of observed point cloud data set from initial time to time $t$, where $p_{i}^{gt}, p_{i}^{s} \in \mathbb{R}^3$ are a single point cloud data and $M$ is the number of point clouds computed from observation $O_t$ depth image. Then the ground truth point cloud covered set by sensor at time $t$ is defined as
\begin{align}
    \mathcal{O}^{c}_{t} := \{p^{gt}_i: ||p^{gt}_i - p^{s}_i||_{2} \leq \delta, p^{gt}_i \in \mathcal{O}^{gt} \mbox{ and } p^{s}_i \in \mathcal{O}_t^s\}
\end{align}
Finally, the surface IoU at time $t$ is $\text{IoU}_t:= \frac{|\mathcal{O}^{c}_{t}|}{|\mathcal{O}^{gt}|}$. Here, we used $\delta=5$ mm.

\paragraph{Chamfer-$L1$ Distance} Another metric we used to evaluate our model is Chamfer-L1 distance~\cite{Mescheder_2019_CVPR}. We define the Chamfer-$L1$ distance $C_t$ between the two 3D point cloud set $\mathcal{O}^{gt}$ and $\mathcal{O}^{s}_{t}$ at time $t$ is defined as follows:
\begin{align}
    C_t &:= \frac{1}{2|\mathcal{O}^{gt}|} \sum_{p^{gt} \in \mathcal{O}^{gt}} \min_{p^{s}\in \mathcal{O}^{s}_{t}} ||p^{s}-p^{gt}|| \nonumber\\
    &+ \frac{1}{2|\mathcal{O}^{s}_{t}|} \sum_{p^{s} \in \mathcal{O}^{s}_{t}} \min_{p^{gt}\in \mathcal{O}^{gt}} ||p^{s}-p^{gt}||
\end{align}

\begin{figure*}[t]
    \centering
    \includegraphics[width=0.8\textwidth]{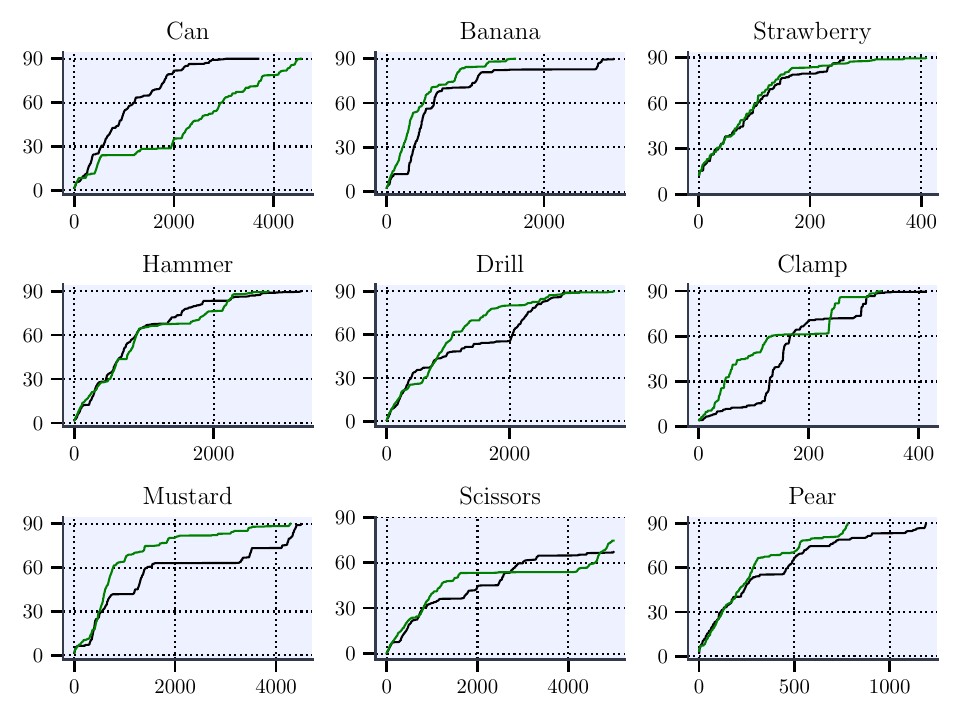}
    \captionsetup{width=.9\textwidth}
    \caption{\textbf{IoU-Step graph} includes AMB-TTA (in black) and AMB-TTS (in green) models, both reaching either 90 \% IoU or 5,000 steps. The horizontal axis represents the number of steps, and the vertical corresponds to the IoU. Small objects like strawberry, achieve 90 \% IoU comparably faster than large objects like can.}
    \label{fig:bar_graph}
\end{figure*}

\begin{figure*}[t]
    \centering
    \includegraphics[width=\textwidth]{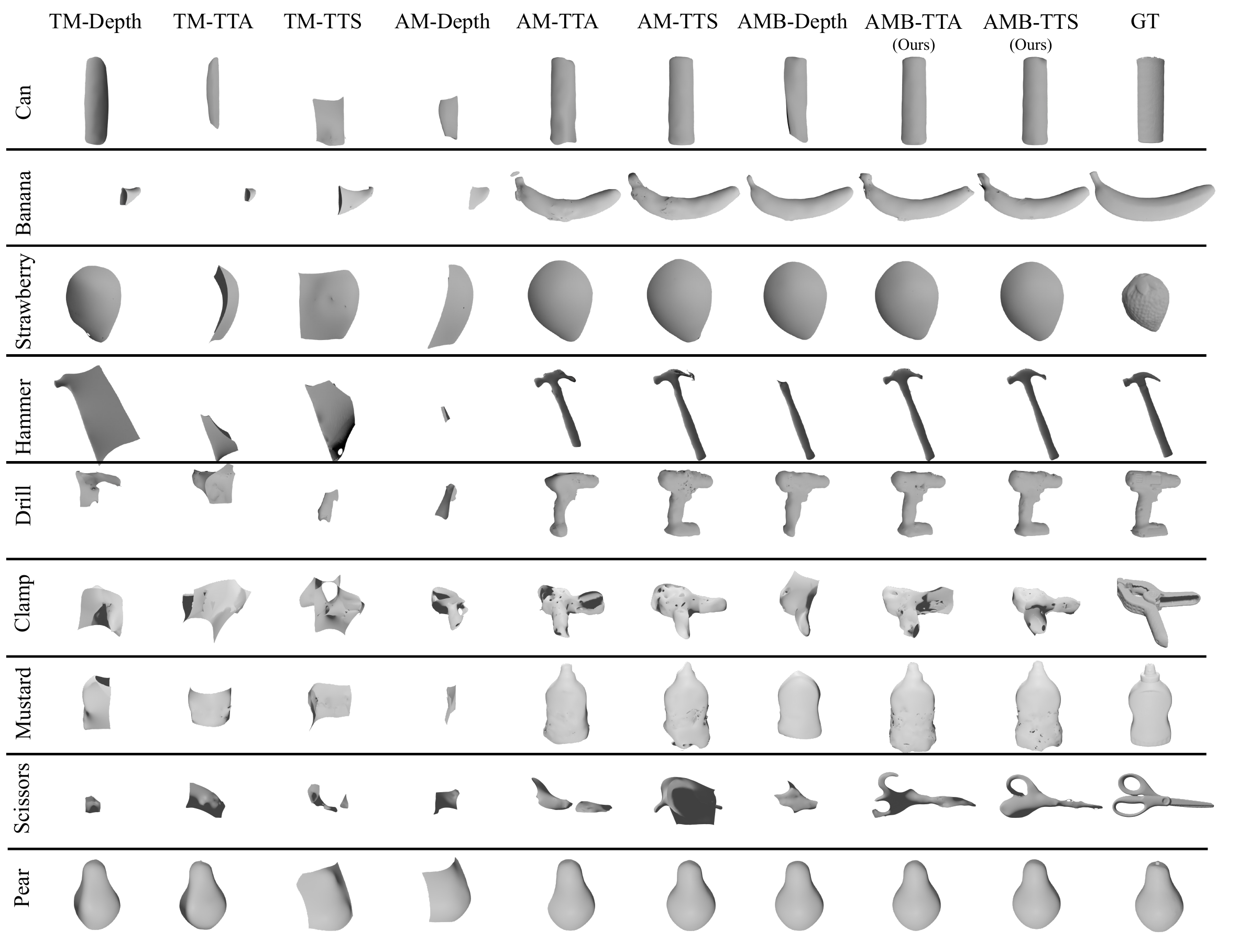}
    \caption{\textbf{Further qualitative results on unseen YCB objects} with different state and reward settings. From active tactile exploration, we obtain point cloud data of tactile depth readings on the object's surface. To generate mesh, we apply Poisson surface reconstruction algorithm~\cite{kazhdan2006poisson}.}
    \label{fig:mesh_result_more}
\end{figure*}
\section{Experiments}
\subsection{Ablation Study}
We ablated the training performance of various primitives shapes on AMB-TTA model, as depicted in Fig.~\ref{fig:ablation}. The Cube-only model exhibited unstable IoU. Conversely, both the Cube-only and Cube + Sphere models showed early stabilization in terms of IoU. Moreover, the Cube + Sphere training model demonstrated a shorter average length, while maintaining a 90 \% IoU, implying a more effective exploration of the objects within fewer steps during training which means having Cube+Sphere results in better generalization even for exploring the training objects.

\subsection{Simulation Environment}
We evaluated the AcTExplore on various YCB objects after training on primitive objects. Fig. \ref{fig:texture} illustrates the diversity of shapes and textures of training and testing environments.

\begin{table*}[ht!]
    \centering
    \renewcommand{\arraystretch}{1.5}
    \rotatebox{90}{
    \begin{minipage}{0.9\textheight}
    \caption{\textbf{Further Quantitative results on unseen YCB objects:} The table presents IoU and Chamfer-L1 distance (cm) \cite{Mescheder_2019_CVPR} values of the predicted meshes and ground-truth meshes. The results were obtained within 5,000 steps. The surface area is listed below each object's name. Lastly, since a recurrent structure is an alternative approach to process temporal information, we implement a PPO variant with LSTM modules to compare with our proposed temporal state representations (TTA/TTS)}
    \label{tab:evaluation}
    \begin{tabular}{l|l||c|c|c|c|c|c|c|c|c}
    \hline\hline
        \multicolumn{2}{c||}{\multirow{2}{*}{\backslashbox[90pt]{Models}{Objects}}} & Can & Banana & Strawberry & Hammer & Drill & Clamp & Mustard & Scissors & Pear \\ \multicolumn{2}{c||}{} &(616 \small{$\text{cm}^2$}) &(216 \small{$\text{cm}^2$}) &(68 \small{$\text{cm}^2$}) &(410 \small{$\text{cm}^2$}) &(591 \small{$\text{cm}^2$}) &(111.13 \small{$\text{cm}^2$}) &( 454.54 \small{$\text{cm}^2$}) &(165.48 \small{$\text{cm}^2$}) &( 172.47\small{$\text{cm}^2$}) \\ \hline
        \multicolumn{7}{c}{} IoU $\uparrow$ \hspace{2 em}(Chamfer-$L_1$ $\downarrow$) \\
        \Xhline{2.5\arrayrulewidth}
        \multirow{3}{*}{TM} & depth   & 31.93 (2.66) & 11.11 (7.52) & 83.60 (0.44) & 32.78 (1.86)& 19.19 (4.1) & 81.76 (1.5) & 10.07 (4.07) & 24.29 (8.15) & 70.95 (0.55) \\ 
         \cline{2-11}
        & TTA  & 17.60 (3.57) & 6.03 (9.03) & 41.0 (1.23) & 14.85 (6.94)& 28.15 (3.99)  &86.29 (1.4) & 19.94 (3.22) & 14.17 (4.98) & 67.89 (0.62) \\ 
         \cline{2-11}
         & TTS & 15.93 (5.22) & 18.23 (5.48) & 57.89 (0.88) & 28.66 (2.47)&15.5 (3.97)  & 58.53 (1.08) & 14.55(2.95) & 11.26(4.97) & 30.13(2.13) \\ 
         \Xhline{2.5\arrayrulewidth}

        \multirow{3}{*}{AM} &depth &11.59 (5.49) &10.22 (6.84)  &47.33 (1.16)  & 5.07 (7.69) &9.49 (4.03)  & 58.81(0.82) & 11.04(5.16) & 5.11(6.78) & 28.75(2.08)\\ 
         \cline{2-11}
        & TTA  & 72.70 (0.56) & 97.70 (0.35) & \textbf{100} (0.28) & 79.80 (0.82) & 57.58 (1.43)  & \textbf{100} (0.62) &71.72 (0.80) & 41.77 (2.87) & 86.07(0.43) \\ 
         \cline{2-11}
         & TTS  & \textbf{98.25} (0.22) & \textbf{100} (0.34) & \textbf{100} (0.31) & 88.22 (0.44)& 99.02 (0.37 )  & \textbf{100} (0.69) & 87.13 (0.59) & 28.37 (2.38) & \textbf{100} (0.23) \\ 
         \Xhline{2.5\arrayrulewidth}
    
        \multirow{4}{*}{AMB} & depth   & 41.45 (1.42) & 98.64 (\textbf{0.25}) & \textbf{100} \textbf{(0.23)} & 61.42 (1.17) & 79.68 (0.95)  & 44.76 (1.77) & 65.74 (0.9)  &31.99 (3.2)  & \textbf{100} (\textbf{0.2}) \\ 
         \cline{2-11}
          & depth+LSTM  & 88.54 (0.3) & 99.96 (0.28) & \textbf{100} (0.24) & 87.54 (0.49) & 92.81 (0.36) & 99.55 (\textbf{0.56}) & 88.33 (0.36) & 29.83 (0.58) & \textbf{100} (\textbf{0.2}) \\
          \cline{2-11}
        & TTA (ours) & 89.6 (0.29)  & \textbf{100} (0.33) & \textbf{100} (0.25) & \textbf{98.22} (0.29)& 98.85 (0.32)  & \textbf{100} (0.66) & \textbf{95.91} (0.51) & 67.02 (0.87) & \textbf{100} (0.22) \\ 
         \cline{2-11}
         & TTS (ours) & 97.45 \textbf{(0.20)} & \textbf{100} (0.3) & \textbf{100} (0.25) & 96.96 \textbf{(0.28)} & \textbf{99.74 (0.31)}  & \textbf{100} (0.59) & 95.02 (\textbf{0.49}) & \textbf{74.62} (\textbf{0.61}) & \textbf{100} (\textbf{0.2}) \\ 
         \hline\hline
    \end{tabular}
    \end{minipage}
    }
\end{table*}



%% file: ms.bbl
\begin{thebibliography}{10}

\bibitem{soft-bubble}
Alex Alspach, Kunimatsu Hashimoto, Naveen Kuppuswarny, and Russ Tedrake.
\newblock Soft-bubble: A highly compliant dense geometry tactile sensor for robot manipulation.
\newblock pages 597--604, 04 2019.

\bibitem{doi:10.1177/0278364919887447}
OpenAI:~Marcin Andrychowicz, Bowen Baker, Maciek Chociej, Rafal Józefowicz, Bob McGrew, Jakub Pachocki, Arthur Petron, Matthias Plappert, Glenn Powell, Alex Ray, Jonas Schneider, Szymon Sidor, Josh Tobin, Peter Welinder, Lilian Weng, and Wojciech Zaremba.
\newblock Learning dexterous in-hand manipulation.
\newblock {\em The International Journal of Robotics Research}, 39(1):3--20, 2020.

\bibitem{azar2017minimax}
Mohammad~Gheshlaghi Azar, Ian Osband, and R{\'e}mi Munos.
\newblock Minimax regret bounds for reinforcement learning.
\newblock In {\em International Conference on Machine Learning}, 2017.

\bibitem{Tac2Pose}
Maria Bauza, Antonia Bronars, and Alberto Rodriguez.
\newblock {Tac2Pose}: {Tactile} {Object} {Pose} {Estimation} from the {First} {Touch}.
\newblock 2022.
\newblock Publisher: arXiv Version Number: 2.

\bibitem{doi:10.1098/rsos.181563}
Wouter~M. Bergmann~Tiest and Astrid M.~L. Kappers.
\newblock The influence of visual and haptic material information on early grasping force.
\newblock {\em Royal Society Open Science}, 6(3):181563, 2019.

\bibitem{trends}
Aude Billard and Danica Kragic.
\newblock Trends and challenges in robot manipulation.
\newblock {\em Science}, 364:eaat8414, 06 2019.

\bibitem{botteghiREINFORCEMENTLEARNINGHELPS2020}
N.~Botteghi, Beril Sirmacek, R.~Schulte, M.~Poel, and C.~Brune.
\newblock {REINFORCEMENT} {LEARNING} {HELPS} {SLAM}: {LEARNING} {TO} {BUILD} {MAPS}.
\newblock {\em International Archives of the Photogrammetry, Remote Sensing \& Spatial Information Sciences}, 43, 2020.

\bibitem{curiosity}
Nicolò Botteghi, Rob Schulte, Beril Sirmacek, Mannes Poel, and Christoph Brune.
\newblock Curiosity-{Driven} {Reinforcement} {Learning} {Agent} for {Mapping} {Unknown} {Indoor} {Environments}.
\newblock {\em ISPRS Annals of the Photogrammetry, Remote Sensing and Spatial Information Sciences}, 1:129--136, 2021.
\newblock Publisher: Copernicus GmbH.

\bibitem{caddeo2023sim2real}
Gabriele~M. Caddeo, Andrea Maracani, Paolo~D. Alfano, Nicola~A. Piga, Lorenzo Rosasco, and Lorenzo Natale.
\newblock Sim2real bilevel adaptation for object surface classification using vision-based tactile sensors, 2023.

\bibitem{10160359}
Gabriele~M. Caddeo, Nicola~A. Piga, Fabrizio Bottarel, and Lorenzo Natale.
\newblock Collision-aware in-hand 6d object pose estimation using multiple vision-based tactile sensors.
\newblock In {\em 2023 IEEE International Conference on Robotics and Automation (ICRA)}, pages 719--725, 2023.

\bibitem{DBLP:journals/corr/CalliWSSAD15}
Berk {\c{C}}alli, Aaron Walsman, Arjun Singh, Siddhartha~S. Srinivasa, Pieter Abbeel, and Aaron~M. Dollar.
\newblock Benchmarking in manipulation research: The {YCB} object and model set and benchmarking protocols.
\newblock {\em CoRR}, abs/1502.03143, 2015.

\bibitem{cappeSelectiveIntegrationAuditoryvisual2009}
Céline Cappe, Gregor Thut, Vincenzo Romei, and Micah~M. Murray.
\newblock Selective integration of auditory-visual looming cues by humans.
\newblock {\em Neuropsychologia}, 47(4):1045--1052, March 2009.

\bibitem{chaplotLearningExploreUsing2020}
Devendra~Singh Chaplot, Dhiraj Gandhi, Saurabh Gupta, Abhinav Gupta, and Ruslan Salakhutdinov.
\newblock Learning to {Explore} using {Active} {Neural} {SLAM}, April 2020.
\newblock arXiv:2004.05155 [cs].

\bibitem{chenSelflearningExplorationMapping2019}
Fanfei Chen, Shi Bai, Tixiao Shan, and Brendan Englot.
\newblock Self-learning exploration and mapping for mobile robots via deep reinforcement learning.
\newblock In {\em Aiaa scitech 2019 forum}, page 0396, 2019.

\bibitem{chenAutonomousExplorationUncertainty2020}
Fanfei Chen, John~D. Martin, Yewei Huang, Jinkun Wang, and Brendan Englot.
\newblock Autonomous exploration under uncertainty via deep reinforcement learning on graphs.
\newblock In {\em 2020 {IEEE}/{RSJ} {International} {Conference} on {Intelligent} {Robots} and {Systems} ({IROS})}, pages 6140--6147. IEEE, 2020.

\bibitem{chenLearningExplorationPolicies2019}
Tao Chen, Saurabh Gupta, and Abhinav Gupta.
\newblock Learning {Exploration} {Policies} for {Navigation}, March 2019.
\newblock arXiv:1903.01959 [cs].

\bibitem{Comi2023TouchSDFAD}
Mauro Comi, Yijiong Lin, Alex Church, Alessio Tonioni, Laurence Aitchison, and Nathan~F. Lepora.
\newblock Touchsdf: A deepsdf approach for 3d shape reconstruction using vision-based tactile sensing.
\newblock {\em ArXiv}, abs/2311.12602, 2023.

\bibitem{DBLP:journals/corr/abs-2103-00655}
Cristiana de~Farias, Naresh Marturi, Rustam Stolkin, and Yasemin Bekiroglu.
\newblock Simultaneous tactile exploration and grasp refinement for unknown objects.
\newblock {\em CoRR}, abs/2103.00655, 2021.

\bibitem{DBLP:journals/corr/abs-2201-01367}
Won~Kyung Do and Monroe Kennedy.
\newblock {DenseTact}: {Optical} {Tactile} {Sensor} for {Dense} {Shape} {Reconstruction}.
\newblock {\em 2022 International Conference on Robotics and Automation (ICRA)}, pages 6188--6194, May 2022.

\bibitem{8793773}
Danny Driess, Daniel Hennes, and Marc Toussaint.
\newblock Active multi-contact continuous tactile exploration with gaussian process differential entropy.
\newblock In {\em 2019 International Conference on Robotics and Automation (ICRA)}, pages 7844--7850, 2019.

\bibitem{pybullet}
Benjamin Ellenberger.
\newblock Pybullet gymperium.
\newblock \url{ https://github.com/benelot/pybullet-gym}, 2018--2019.

\bibitem{ganguly2022gradtac}
Kanishka Ganguly, Pavan Mantripragada, Chethan~M Parameshwara, Cornelia Ferm{\"u}ller, Nitin~J Sanket, and Yiannis Aloimonos.
\newblock Gradtac: Spatio-temporal gradient based tactile sensing.
\newblock {\em Frontiers in Robotics and AI}, 9:898075, 2022.

\bibitem{ganguly2020grasping2}
Kanishka Ganguly, Behzad Sadrfaridpour, Krishna~Bhavithavya Kidambi, Cornelia Ferm{\"u}ller, and Yiannis Aloimonos.
\newblock Grasping in the dark: Compliant grasping using shadow dexterous hand and biotac tactile sensor.

\bibitem{DBLP:journals/corr/abs-1712-03316}
Daniel Gordon, Aniruddha Kembhavi, Mohammad Rastegari, Joseph Redmon, Dieter Fox, and Ali Farhadi.
\newblock {IQA:} visual question answering in interactive environments.
\newblock {\em CoRR}, abs/1712.03316, 2017.

\bibitem{hoganSeeingYourSkin2021}
Francois~Robert Hogan, Michael Jenkin, Sahand Rezaei{-}Shoshtari, Yogesh~A. Girdhar, David Meger, and Gregory Dudek.
\newblock Seeing through your skin: Recognizing objects with a novel visuotactile sensor.
\newblock volume abs/2011.09552, 2020.

\bibitem{jin2018qlearning}
Chi Jin, Zeyuan Allen-Zhu, Sebastien Bubeck, and Michael~I Jordan.
\newblock Is q-learning provably efficient?
\newblock In S.~Bengio, H.~Wallach, H.~Larochelle, K.~Grauman, N.~Cesa-Bianchi, and R.~Garnett, editors, {\em Advances in Neural Information Processing Systems}, volume~31. Curran Associates, Inc., 2018.

\bibitem{JOHANSSON198217}
R.S. Johansson, U.~Landstro¨m, and R.~Lundstro¨m.
\newblock Responses of mechanoreceptive afferent units in the glabrous skin of the human hand to sinusoidal skin displacements.
\newblock {\em Brain Research}, 244(1):17--25, 1982.

\bibitem{doi:10.1098/rstb.2011.0171}
Astrid M.~L. Kappers.
\newblock Human perception of shape from touch.
\newblock {\em Philosophical Transactions of the Royal Society B: Biological Sciences}, 366(1581):3106--3114, 2011.

\bibitem{KATUS2015275}
Tobias Katus and Søren~K. Andersen.
\newblock Chapter 21 - the role of spatial attention in tactile short-term memory.
\newblock In Pierre Jolicoeur, Christine Lefebvre, and Julio Martinez-Trujillo, editors, {\em Mechanisms of Sensory Working Memory}, pages 275--292. Academic Press, San Diego, 2015.

\bibitem{kazhdan2006poisson}
Michael Kazhdan, Matthew Bolitho, and Hugues Hoppe.
\newblock Poisson surface reconstruction.
\newblock In {\em Proceedings of the fourth Eurographics symposium on Geometry processing}, volume~7, page~0, 2006.

\bibitem{kelestemurTactilePoseEstimation2022}
Tarik Kelestemur, Robert Platt, and Taskin Padir.
\newblock Tactile {Pose} {Estimation} and {Policy} {Learning} for {Unknown} {Object} {Manipulation}.
\newblock 2022.
\newblock Publisher: arXiv Version Number: 1.

\bibitem{DBLP:journals/corr/abs-2005-14679}
Mike Lambeta, Po{-}Wei Chou, Stephen Tian, Brian~H. Yang, Benjamin Maloon, Victoria~Rose Most, Dave Stroud, Raymond Santos, Ahmad Byagowi, Gregg Kammerer, Dinesh Jayaraman, and Roberto Calandra.
\newblock {DIGIT:} {A} novel design for a low-cost compact high-resolution tactile sensor with application to in-hand manipulation.
\newblock {\em CoRR}, abs/2005.14679, 2020.

\bibitem{LAWSON2015239}
Rebecca Lawson, Alexandra~M. Fernandes, Pedro~B. Albuquerque, and Simon Lacey.
\newblock Chapter 19 - remembering touch: Using interference tasks to study tactile and haptic memory.
\newblock In Pierre Jolicoeur, Christine Lefebvre, and Julio Martinez-Trujillo, editors, {\em Mechanisms of Sensory Working Memory}, pages 239--259. Academic Press, San Diego, 2015.

\bibitem{liang2019salientdso}
Huai-Jen Liang, Nitin~J Sanket, Cornelia Ferm{\"u}ller, and Yiannis Aloimonos.
\newblock Salientdso: Bringing attention to direct sparse odometry.
\newblock {\em IEEE Transactions on Automation Science and Engineering}, 16(4):1619--1626, 2019.

\bibitem{9562061}
Wenyu Liang, Qinyuan Ren, Xiaoqiao Chen, Junli Gao, and Yan Wu.
\newblock Dexterous manoeuvre through touch in a cluttered scene.
\newblock In {\em 2021 IEEE International Conference on Robotics and Automation (ICRA)}, pages 6308--6314, 2021.

\bibitem{linDTactVisionBasedTactile2022}
Changyi Lin, Ziqi Lin, Shaoxiong Wang, and Huazhe Xu.
\newblock {DTact}: {A} {Vision}-{Based} {Tactile} {Sensor} that {Measures} {High}-{Resolution} {3D} {Geometry} {Directly} from {Darkness}.
\newblock 2022.
\newblock Publisher: arXiv Version Number: 1.

\bibitem{linLearningIdentifyObject2019}
Justin Lin, Roberto Calandra, and Sergey Levine.
\newblock Learning to {Identify} {Object} {Instances} by {Touch}: {Tactile} {Recognition} via {Multimodal} {Matching}.
\newblock In {\em 2019 {International} {Conference} on {Robotics} and {Automation} ({ICRA})}, pages 3644--3650, May 2019.
\newblock ISSN: 2577-087X.

\bibitem{luCuriosityDrivenSelfsupervised2022}
Yujie Lu, Jianren Wang, and Vikash Kumar.
\newblock Curiosity {Driven} {Self}-supervised {Tactile} {Exploration} of {Unknown} {Objects}.
\newblock 2022.
\newblock Publisher: arXiv Version Number: 1.

\bibitem{SLAMReview}
Andréa Macario~Barros, Maugan Michel, Yoann Moline, Gwenolé Corre, and Frédérick Carrel.
\newblock A comprehensive survey of visual slam algorithms.
\newblock {\em Robotics}, 11(1), 2022.

\bibitem{7177729}
Uriel Martinez-Hernandez, Nathan~F. Lepora, and Tony~J. Prescott.
\newblock Active haptic shape recognition by intrinsic motivation with a robot hand.
\newblock In {\em 2015 IEEE World Haptics Conference (WHC)}, pages 299--304, 2015.

\bibitem{Massari}
Luca Massari, Calogero Oddo, Edoardo Sinibaldi, Renaud Detry, Joseph Bowkett, and Kalind Carpenter.
\newblock Tactile sensing and control of robotic manipulator integrating fiber bragg grating strain-sensor.
\newblock {\em Frontiers in Neurorobotics}, 13, 04 2019.

\bibitem{Mescheder_2019_CVPR}
Lars Mescheder, Michael Oechsle, Michael Niemeyer, Sebastian Nowozin, and Andreas Geiger.
\newblock Occupancy networks: Learning 3d reconstruction in function space.
\newblock In {\em Proceedings of the IEEE/CVF Conference on Computer Vision and Pattern Recognition (CVPR)}, June 2019.

\bibitem{DBLP:journals/corr/MirowskiPVSBBDG16}
Piotr Mirowski, Razvan Pascanu, Fabio Viola, Hubert Soyer, Andrew~J. Ballard, Andrea Banino, Misha Denil, Ross Goroshin, Laurent Sifre, Koray Kavukcuoglu, Dharshan Kumaran, and Raia Hadsell.
\newblock Learning to navigate in complex environments.
\newblock {\em CoRR}, abs/1611.03673, 2016.

\bibitem{munozMultisensoryPerceptionUncertain2012}
Nicole~E. Munoz and Daniel~T. Blumstein.
\newblock Multisensory perception in uncertain environments.
\newblock {\em Behavioral Ecology}, 23(3):457--462, May 2012.

\bibitem{DBLP:journals/corr/abs-2103-16747}
Yashraj Narang*, Balakumar Sundaralingam*, Miles Macklin, Arsalan Mousavian, and Dieter Fox.
\newblock Sim-to-real for robotic tactile sensing via physics-based simulation and learned latent projections (*equal contribution).
\newblock {\em IEEE Intl. Conf. on Robotics and Automation}, pages 6444--6451, 2021.

\bibitem{DBLP:journals/corr/abs-2101-05452}
Yashraj~S. Narang, Balakumar Sundaralingam, Karl~Van Wyk, Arsalan Mousavian, and Dieter Fox.
\newblock Interpreting and predicting tactile signals for the syntouch biotac.
\newblock {\em CoRR}, abs/2101.05452, 2021.

\bibitem{OttenhausActiveExploration}
Simon Ottenhaus, Lukas Kaul, Nikolaus Vahrenkamp, and Tamim Asfour.
\newblock Active tactile exploration based on cost-aware information gain maximization.
\newblock {\em International Journal of Humanoid Robotics}, 15:1850015, 02 2018.

\bibitem{parkBiomimeticElastomericRobot2022}
K.~Park, H.~Yuk, M.~Yang, J.~Cho, H.~Lee, and J.~Kim.
\newblock A biomimetic elastomeric robot skin using electrical impedance and acoustic tomography for tactile sensing.
\newblock {\em Science Robotics}, 7(67):eabm7187, June 2022.
\newblock Publisher: American Association for the Advancement of Science.

\bibitem{pecynaVisualTactileMultimodalityFollowing2022}
Leszek Pecyna, Siyuan Dong, and Shan Luo.
\newblock Visual-{Tactile} {Multimodality} for {Following} {Deformable} {Linear} {Objects} {Using} {Reinforcement} {Learning}.
\newblock {\em 2022 IEEE/RSJ International Conference on Intelligent Robots and Systems (IROS)}, pages 3987--3994, October 2022.
\newblock Conference Name: 2022 IEEE/RSJ International Conference on Intelligent Robots and Systems (IROS) ISBN: 9781665479271 Place: Kyoto, Japan Publisher: IEEE.

\bibitem{placedDeepReinforcementLearning2020}
Julio~A. Placed and José~A. Castellanos.
\newblock A {Deep} {Reinforcement} {Learning} {Approach} for {Active} {SLAM}.
\newblock {\em Applied Sciences}, 10(23):8386, January 2020.
\newblock Number: 23 Publisher: Multidisciplinary Digital Publishing Institute.

\bibitem{ActiveSLAMReview}
Julio~A. Placed, Jared Strader, Henry Carrillo, Nikolay Atanasov, Vadim Indelman, Luca Carlone, and José~A. Castellanos.
\newblock A {Survey} on {Active} {Simultaneous} {Localization} and {Mapping}: {State} of the {Art} and {New} {Frontiers}.
\newblock {\em IEEE Transactions on Robotics}, pages 1--20, 2023.
\newblock Conference Name: IEEE Transactions on Robotics.

\bibitem{10410896}
Parth Potdar, David Hardman, Elijah Almanzor, and Fumiya Iida.
\newblock High-speed tactile braille reading via biomimetic sliding interactions.
\newblock {\em IEEE Robotics and Automation Letters}, 9(3):2614--2621, 2024.

\bibitem{stable-baselines3}
Antonin Raffin, Ashley Hill, Adam Gleave, Anssi Kanervisto, Maximilian Ernestus, and Noah Dormann.
\newblock Stable-baselines3: Reliable reinforcement learning implementations.
\newblock {\em Journal of Machine Learning Research}, 22(268):1--8, 2021.

\bibitem{Ramani2019ASS}
Dhruv Ramani.
\newblock A short survey on memory based reinforcement learning.
\newblock {\em ArXiv}, abs/1904.06736, 2019.

\bibitem{DBLP:journals/corr/abs-2007-03778}
Edward~J. Smith, Roberto Calandra, Adriana Romero, Georgia Gkioxari, David Meger, Jitendra Malik, and Michal Drozdzal.
\newblock 3d shape reconstruction from vision and touch.
\newblock {\em CoRR}, abs/2007.03778, 2020.

\bibitem{Suresh22icra}
S.~Suresh, Z.~Si, J.~Mangelson, W.~Yuan, and M.~Kaess.
\newblock {ShapeMap 3-D}: Efficient shape mapping through dense touch and vision.
\newblock In {\em Proc. IEEE Intl. Conf. on Robotics and Automation, ICRA}, Philadelphia, PA, USA, May 2022.

\bibitem{sureshTactileSLAMRealtime2021}
Sudharshan Suresh, Maria Bauza, Kuan-Ting Yu, Joshua~G. Mangelson, Alberto Rodriguez, and Michael Kaess.
\newblock Tactile {SLAM}: {Real}-time inference of shape and pose from planar pushing.
\newblock In {\em 2021 {IEEE} {International} {Conference} on {Robotics} and {Automation} ({ICRA})}, pages 11322--11328, May 2021.
\newblock ISSN: 2577-087X.

\bibitem{suresh2023neuralfeels}
Sudharshan Suresh, Haozhi Qi, Tingfan Wu, Taosha Fan, Luis Pineda, Mike Lambeta, Jitendra Malik, Mrinal Kalakrishnan, Roberto Calandra, Michael Kaess, Joseph Ortiz, and Mustafa Mukadam.
\newblock {N}eural feels with neural fields: {V}isuo-tactile perception for in-hand manipulation.
\newblock In {\em arXiv preprint arXiv:2312.1346}, December 2023.

\bibitem{suresh2022midastouch}
Sudharshan Suresh, Zilin Si, Stuart Anderson, Michael Kaess, and Mustafa Mukadam.
\newblock {M}idas{T}ouch: {M}onte-{C}arlo inference over distributions across sliding touch.
\newblock In {\em Proc. Conf. on Robot Learning, CoRL}, Auckland, NZ, December 2022.

\bibitem{biotac}
SynTouch.
\newblock Biotac product manual.
\newblock Aug 2018.

\bibitem{DBLP:journals/corr/abs-2012-08456}
Shaoxiong Wang, Mike Lambeta, Po{-}Wei Chou, and Roberto Calandra.
\newblock {TACTO:} {A} fast, flexible and open-source simulator for high-resolution vision-based tactile sensors.
\newblock {\em CoRR}, abs/2012.08456, 2020.

\bibitem{https://doi.org/10.1002/adma.202203073}
Xuelian Wei, Baocheng Wang, Zhiyi Wu, and Zhong~Lin Wang.
\newblock An open-environment tactile sensing system: Toward simple and efficient material identification.
\newblock {\em Advanced Materials}, 34(29):2203073, 2022.

\bibitem{xu2023tandem3d}
Jingxi Xu, Han Lin, Shuran Song, and Matei~T. Ciocarlie.
\newblock Tandem3d: Active tactile exploration for 3d object recognition, 2022.

\bibitem{7759723}
Zhengkun Yi, Roberto Calandra, Filipe Veiga, Herke van Hoof, Tucker Hermans, Yilei Zhang, and Jan Peters.
\newblock Active tactile object exploration with gaussian processes.
\newblock In {\em 2016 IEEE/RSJ International Conference on Intelligent Robots and Systems (IROS)}, pages 4925--4930, 2016.

\bibitem{s17122762}
Wenzhen Yuan, Siyuan Dong, and Edward~H. Adelson.
\newblock Gelsight: High-resolution robot tactile sensors for estimating geometry and force.
\newblock {\em Sensors}, 17(12), 2017.

\bibitem{yuan2023robot}
Ying Yuan, Haichuan Che, Yuzhe Qin, Binghao Huang, Zhao-Heng Yin, Kang-Won Lee, Yi~Wu, Soo-Chul Lim, and Xiaolong Wang.
\newblock Robot synesthesia: In-hand manipulation with visuotactile sensing, 2023.

\bibitem{zhao2023fingerslam}
Jialiang Zhao, Maria Bauz{\'a}, and Edward~H. Adelson.
\newblock Fingerslam: Closed-loop unknown object localization and reconstruction from visuo-tactile feedback, 2023.

\end{thebibliography}
